%% file: main.tex
\documentclass[letterpaper]{article}
\usepackage[T1]{fontenc}
\usepackage{textcomp}
\usepackage[latin9]{inputenc}
\usepackage{color}
\usepackage{colortbl}
\usepackage{array}
\usepackage{booktabs}
\usepackage{multirow}
\usepackage{varwidth}
\usepackage{amsmath}
\usepackage{amsthm}
\usepackage{amssymb}
\usepackage{graphicx}

\makeatletter

\providecommand{\tabularnewline}{\\}
\newenvironment{cellvarwidth}[1][t]
    {\begin{varwidth}[#1]{\linewidth}}
    {\@finalstrut\@arstrutbox\end{varwidth}}

\usepackage[preprint]{aaai2027}  
\usepackage[hyphens]{url}  
\usepackage{graphicx} 
\usepackage{natbib}  
\usepackage{caption} 
\usepackage{algorithm}
\usepackage{algorithmic}

\usepackage{newfloat}
\usepackage{listings}
\DeclareCaptionStyle{ruled}{labelfont=normalfont,labelsep=colon,strut=off} 
\floatstyle{ruled}
\newfloat{listing}{tb}{lst}{}
\floatname{listing}{Listing}

\usepackage{booktabs}

\title{Modaliff: A Unified Token Diffusion Model for Any-to-Any Generation}
\author{
	Duc Kieu\textsuperscript{\rm 1}, 
	Kien Do\textsuperscript{\rm 2}, 
	Minh Khoa Le\textsuperscript{\rm 1}, 
	Tung Kieu\textsuperscript{\rm 3}, 
	Dang Nguyen\textsuperscript{\rm 1}, 
	Thin Nguyen\textsuperscript{\rm 1}
}
\affiliations{
    \textsuperscript{\rm 1}Applied Artificial Intelligence Initiative, Deakin University\\
    \textsuperscript{\rm 2}VinSmart Future, 
	\textsuperscript{\rm 3}Aalborg University\\
}

\AtBeginDocument{%
  \let\cite\citep
}

\ifdefined\showcaptionsetup
 \PassOptionsToPackage{caption=false}{subfig}
\fi
\usepackage{subfig}
\makeatother

\begin{document}
\maketitle

\global\long\def\Expect{\mathbb{E}}%
\global\long\def\Real{\mathbb{R}}%
\global\long\def\Data{\mathcal{D}}%
\global\long\def\Loss{\mathcal{L}}%
\global\long\def\Normal{\mathcal{N}}%
\global\long\def\softmax{\text{softmax}}%
\global\long\def\ELBO{\text{ELBO}}%
\global\long\def\argmin#1{\underset{#1}{\text{argmin}}}%
\global\long\def\argmax#1{\underset{#1}{\text{argmax}}}%
\global\long\def\Model{\text{Model}}%

\begin{abstract}
\input{abstract.tex}

\end{abstract}

\section{Introduction}

\input{intro.tex}

\section{Background\label{sec:Preliminaries}}

\input{prelim.tex}

\section{Modaliff}

\input{method.tex}

\section{Experiments\label{sec:Experiment}}

\input{exp.tex}

\section{Related Work}

\input{relate.tex}

\section{Conclusion}

\input{discuss.tex}

\clearpage{}

\bibliography{reference}

\clearpage{}

\appendix
\onecolumn\input{appdx.tex}

\end{document}

%% file: abstract.tex
Any-to-any generation seeks a single model that translates flexibly
among text, image, and audio, yet existing approaches either require
fully aligned multimodal tuples or a shared representation that suppresses
modality-specific structure. We introduce Modaliff, a novel token-based
diffusion model that learns from widely available bimodal pairs while
preserving each modality in its native latent sequence. Modaliff assembles
conditions and targets into a unified token sequence, with per-token
attributes to disambiguate token identities and role-based positional
embedding to preserve each modality's internal structure across sequence
layouts. This design enables a single denoiser to support diverse
generation tasks without task-specific architectural changes or a
shared cross-modal latent space. Beyond one-to-one generation, Modaliff
supports multiple conditions either by combining single-condition
predictions at inference or by lightweight finetuning on captions
that jointly describe multiple inputs. Experiments across text, image,
and audio show that Modaliff achieves the best fidelity among any-to-any
models on most generation tasks while using substantially fewer parameters
and training data.

%% file: intro.tex
A long-standing goal of generative modeling is to build a single model
that translates freely among text, image, and audio, such as captioning
a photograph, imagining the scene behind a sound, or synthesizing
audio from a description. Recent diffusion and flow-based models have
brought this any-to-any goal closer~\cite{bao2023one,tang2023any,li2025omniflow},
building on progress in text-to-image~\cite{rombach2022high} and
text-to-audio~\cite{liu2023audioldm} generation. Yet flexible and
efficient any-to-any generation remains challenging because different
modalities have fundamentally different structures, with text is sequential
and symbolic, image is spatially structured, and audio is temporally
continuous.

Existing any-to-any diffusion models typically follow one of two approaches.
The first approach, exemplified by UniDiffuser~\cite{bao2023one}
and OmniFlow~\cite{li2025omniflow}, concatenates tokens from all
modalities and model their joint density over concatenated multimodal
tokens with a single transformer. This design is general, but it often
requires fully aligned multimodal tuples, which are scarce in practice,
and is less flexible for conditional tasks such as Image + Text $\rightarrow$
Image. The second, represented by CoDi~\cite{tang2023any} and FlowBind~\cite{cha2026flowbind},
instead maps all modalities into a shared representation space and
condition generation there. This reduces the tuple requirement, but
it either requires contrastive pretraining or forces all modalities
into the same vector format, limiting representation expressivity.

We introduce Modaliff, a novel token-based diffusion model for any-to-any
generation that learns from widely available bimodal pairs without
requiring fully aligned multimodal tuples or a shared cross-modal
latent space. Our key idea is to represent each modality as a latent
token sequence and assemble condition and target modalities into a
unified token sequence across generation tasks. To this end, each
token of the sequence carries a rich attribute tuple that specifies
its modality (e.g., text, image, or audio), role (source or target),
and noise level. A role-based positional embedding further organizes
token positions across task-varying sequence layouts. Together, these
components allow a single denoiser to support flexible source and
target combinations without architectural modification, while preserving
expressive conditioning through attention over native token sequences.

Beyond one-to-one generation, Modaliff supports multiple conditions
through two complementary mechanisms, namely inference-time score
composition and lightweight caption-based adaptation. Score composition
adopts a product-of-experts formulation to combine the single-condition
predictions learned from paired data, enabling multi-condition generation
without additional training. In contrast, caption-based adaptation
expresses the joint semantics of multiple conditions in language and
finetunes Modaliff on a compact set of examples constructed from existing
paired data, better capturing interactions among conditions.

Extensive experiments across text, image, and audio demonstrate that
Modaliff achieves the best fidelity among any-to-any models on most
generation tasks. Compared to OmniFlow and CoDi, Modaliff uses six
to eight times fewer parameters and about 5\% of OmniFlow's training
data while maintaining competitive results.

Our contributions are as follows:
\begin{itemize}
\item We propose Modaliff, an any-to-any diffusion model that learns from
bimodal paired data while preserving each modality in its native latent
token space. To our knowledge, Modaliff is the first any-to-any model
that requires neither a shared cross-modal representation nor fully
aligned multimodal tuples for training.
\item We develop two approaches for multi-condition generation through inference-time
score composition and lightweight adaptation on caption-based supervision.
\item Across text, image, and audio, Modaliff achieves the best fidelity
among any-to-any models on most generation tasks while using substantially
fewer parameters and less training data than OmniFlow and CoDi.
\end{itemize}

%% file: prelim.tex
\subsection{Diffusion Models}

In diffusion models \cite{sohl2015deep,song2019generative,ho2020denoising,song2020denoising},
a clean sample $x$ is progressively corrupted by Gaussian noise through
a predefined process:
\begin{equation}
x_{t}=a_{t}x+\sigma_{t}\epsilon,\ \ \epsilon\sim\mathcal{N}\left(0,\mathrm{I}\right)\label{eq:diff_fwd_xt}
\end{equation}
where $a_{t}$ decreases from 1 to 0 and $\sigma_{t}$ increases from
0 to 1 as $t$ goes from $0$ to $T$.

The generative (reverse) process starts from $x_{T}\sim\mathcal{N}\left(0,\mathrm{I}\right)$
and iteratively denoises using $p_{\theta}\left(x_{t-1}|x_{t}\right)$
-- a Gaussian transition whose mean depends on the network $\epsilon_{\theta}$
that predicts the noise. The model is trained with the denoising objective:
\begin{equation}
\mathcal{L}_{\text{DM}}\left(\theta\right)=\Expect_{x,t,\epsilon}\left[\left\Vert \epsilon_{\theta}\left(x_{t},t\right)-\epsilon\right\Vert ^{2}_{2}\right]
\end{equation}

For conditional generation, classifier-free guidance (CFG) is usually
applied:
\begin{equation}
\tilde{\epsilon}_{\theta}\left(x_{t},t,y\right)=w\epsilon_{\theta}\left(x_{t},t,y\right)+\left(1-w\right)\epsilon_{\theta}\left(x_{t},t,\varnothing\right)\label{eq:cfg}
\end{equation}
where $\epsilon_{\theta}\left(x_{t},t,y\right)$ and $\epsilon_{\theta}\left(x_{t},t,\varnothing\right)$
are the conditional and unconditional noise predictions, and $w$
controls the guidance.

\subsection{Diffusion Router}

Diffusion Router (DR)~\cite{kieu2026universal} is a variant of diffusion
models designed for the Universal Multi-Domain Translation (UMDT)
problem. Given $K-1$ paired datasets $\mathcal{D}_{k,c}=\left\{ \left(x^{k},x^{c}\right)\right\} ^{N_{k}}_{n=1}$
between a central domain $X^{c}$ and $K-1$ non-central domains $X^{k}$
($k\neq c$), DR trains a unified noise predictor $\epsilon_{\theta}\left(x^{\text{tgt}}_{t},t,x^{\text{src}},\text{tgt},\text{src}\right)$
which supports bidirectional central-noncentral translations $X^{c}\leftrightarrow X^{k}$
using the objective:
\begin{align}
\mathcal{L}_{\text{pair}}\left(\theta\right) & =\mathbb{E}_{\left(x^{k},x^{c}\right)\sim\mathcal{D}_{k,c},t,\epsilon,\zeta}\bigg[\zeta\left\Vert \epsilon_{\theta}\left(x^{k}_{t},t,x^{c},k,c\right)-\epsilon\right\Vert ^{2}_{2}\nonumber \\
 & \ \ \ \ \ \ \ \ +\left(1-\zeta\right)\left\Vert \epsilon_{\theta}\left(x^{c}_{t},t,x^{k},c,k\right)-\epsilon\right\Vert ^{2}_{2}\bigg],\label{eq:paired_loss}
\end{align}

At inference, DR can translate between two non-central domains $X^{i}$,
$X^{j}$ indirectly via the central domain $X^{c}$. To enable direct
translation, an additional objective is introduced:
\begin{align}
\mathcal{L}_{\text{unpair}}\left(\theta\right) & =\mathbb{E}_{x^{j}_{t}\sim p\left(x^{j}_{t}|x^{c}\right)}\left[\left\Vert \epsilon_{\theta}\left(x^{j}_{t},t,x^{i},j,i\right)\right.\right.\nonumber \\
 & \ \ \ \ \ \ \ \ \ \left.\left.-\text{sg}\left(\epsilon_{\theta}\left(x^{j}_{t},t,x^{c},j,c\right)\right)\right\Vert ^{2}_{2}\right],\label{eq:unpaired_loss}
\end{align}

\noindent where $\text{sg}\left(\cdot\right)$ denotes the stop gradient
operator. The final training objective is:
\begin{align}
\Loss_{\text{final}}\left(\theta\right) & =\lambda_{1}\mathcal{L}_{\text{unpair}}\left(\theta\right)+\lambda_{2}\mathcal{L}_{\text{pair}}\left(\theta\right)\label{eq:final_loss}
\end{align}

Despite promising results, DR is primarily designed for image-to-image
translation, assuming all domains share the same dimensionality. This
limits its applicability to multimodal settings, where domains may
differ in representation and structure. In addition, DR focuses on
pairwise translation and does not naturally support composing multiple
conditions for generation.

%% file: method.tex
\begin{figure*}
\begin{centering}
\subfloat[Overall architecture]{\begin{centering}
\includegraphics[totalheight=0.35\textheight]{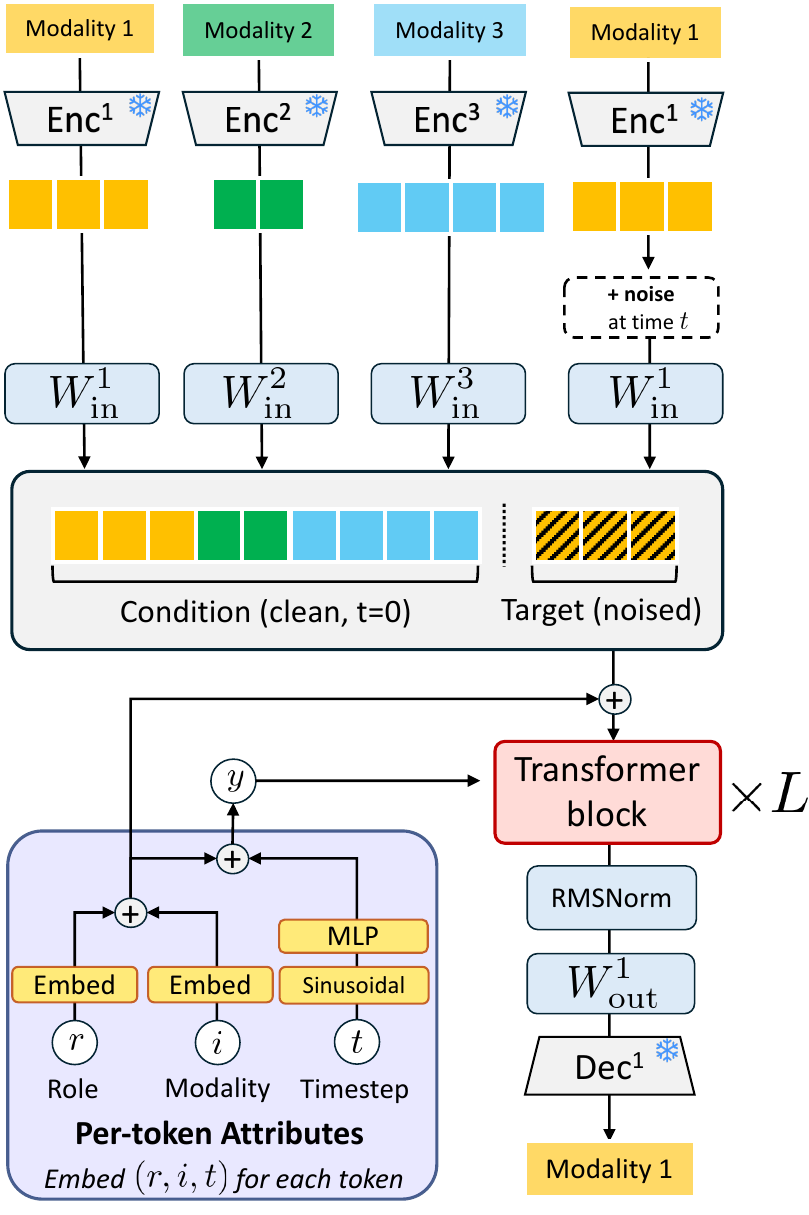}
\par\end{centering}
}\hspace*{\fill}\subfloat[A transformer block]{\begin{centering}
\includegraphics[totalheight=0.35\textheight]{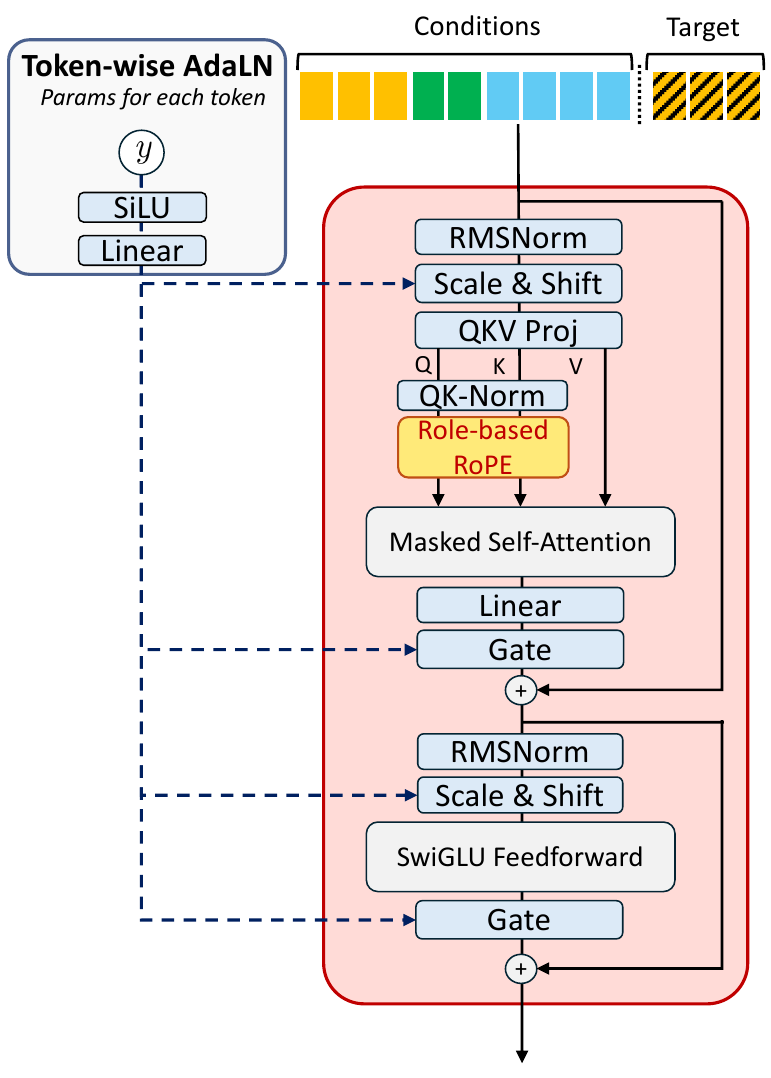}
\par\end{centering}

}\hspace*{\fill}\subfloat[Token attributes and role-based PE.]{\begin{centering}
\includegraphics[totalheight=0.35\textheight]{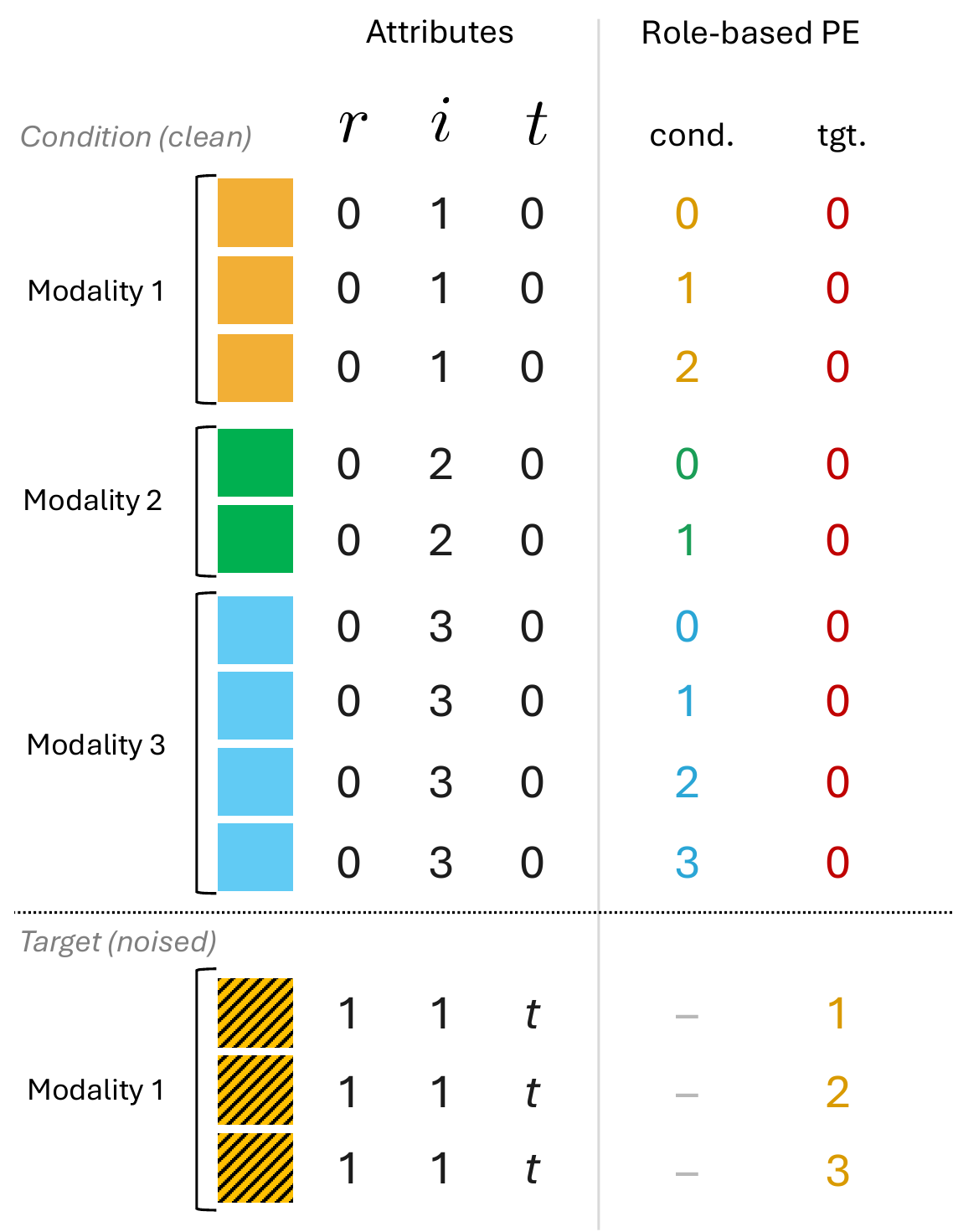}
\par\end{centering}
}
\par\end{centering}
\caption{Fine-grain illustration of our proposed Modaliff.\label{fig:arch} }
\end{figure*}

\subsection{Unified Token-based Representation\label{subsec:unified}}

A fundamental challenge in multimodal generation is that latent representations
of different modalities differ substantially in both structure and
semantic abstraction. Existing unified diffusion and flow models typically
address this heterogeneity either by modeling a joint distribution
over concatenated multimodal tokens \cite{bao2023one,li2025omniflow}
or by mapping modalities into a shared representation space \cite{tang2023any,cha2026flowbind}.
The former often requires fully aligned multimodal datasets \cite{li2025omniflow}
and struggles to support flexible conditioning patterns such as Image
+ Text $\rightarrow$ Image, while the latter typically discards modality-specific
structure and restricts cross-modal interaction to operations between
fixed-dimensional vectors.

To address these limitations, we represent each modality as its native
latent token sequence and concatenate arbitrary condition and target
modalities into a unified sequence. Specifically, let $\mathcal{M}=\left\{ 1,\ldots,K\right\} $
denote the modality set, instantiated with text, image, and audio,
following prior multimodal settings~\cite{cha2026flowbind}. For
each modality $i\in\mathcal{M}$, a frozen encoder $\text{Enc}^{i}\left(\cdot\right)$
encodes a sample $x^{i}$ to a token sequence, $z^{i}=\text{Enc}^{i}\left(x^{i}\right),z^{i}\in\mathbb{R}^{N_{i}\times d_{i}}$,
where $N_{i}$ and $d_{i}$ denote the number and dimensionality of
tokens. Given a generation task, let $\mathcal{C},\mathcal{T}\subseteq\mathcal{M}$
denote the condition and target modality sets, respectively. We concatenate
the token sequences of the context and target set as:
\begin{align}
H_{t}= & \mathcal{H}\left(\left\{ z^{i}\right\} _{i\in\mathcal{C}},\left\{ z^{j}_{t}\right\} _{j\in\mathcal{T}}\right)\nonumber \\
= & \text{Concat}\left(\left\{ z^{i}\cdot W^{i}_{\text{in}}\right\} _{i\in\mathcal{C}},\left\{ z^{j}_{t}\cdot W^{j}_{\text{in}}\right\} _{j\in\mathcal{T}}\right),\label{eq:Ht}
\end{align}

\noindent where $W^{i}_{\text{in}}\in\mathbb{R}^{d_{i}\times d}$
is a trainable projection matrix. Here, each condition modality $i\in\mathcal{C}$
contributes clean sequence $z^{i}$, while each target modality $j\in\mathcal{T}$
contributes a noisy sequence $z^{j}_{t}$ at diffusion timestep $t$,
following Eq.~\ref{eq:diff_fwd_xt}. This design reduces flexible
generation to rearranging condition and target segments, while the
shared network and modality-native representations remain unchanged.

Processing all tasks through a unified sequence introduces two challenges.
First, $H_{t}$ does not specify each token\textquoteright s modality,
condition or target status, or diffusion timestep, leaving the generation
task ambiguous to the shared model. Second, the sequence length and
segment layout vary across tasks, making conventional positional indexing
unsuitable. Global indexing over $H_{t}$ \cite{bao2023one} binds
modalities to a fixed segment order, limiting flexibility, whereas
local indexing avoids this constraint but creates artificial correspondences
between unrelated tokens sharing the same index, introducing false
positional biases into the model.

To resolve the first challenge, we define an attribute tuple $\left(r,i,t\right)$
for each token, encoding its generation role, modality type, and diffusion
timestep. Specifically, $r\in\left\{ 0,1\right\} $, where $r=0$
denotes a condition token and $r=1$ denotes a target token, $i\in\mathcal{M}$
identifies the modality type, and $t$ records its noise level, with
$t=0$ for clean condition tokens and $t$ set to the current denoising
timestep for target tokens. These attributes are embedded and injected
into each transformer block through a token-wise AdaLN~\cite{peebles2023scalable},
which produces token-specific scale, shift, and gating parameters,
enabling the shared network to adapt its computation to each token
without modality-specific branches. 

We address the second challenge with role-based positional embedding
(PE) built on RoPE~\cite{su2024roformer}, in which the positions
used in attention depend on whether the attending token belongs to
a condition or target segment. Condition queries use local positions
and attend only within their own segment, preserving internal structure
and avoiding false correspondences across conditions. Target queries
retain local target positions while treating all condition segments
as an unordered set under a shared positional reference, removing
spurious cross-segment biases and supporting flexible condition-target
layouts.

Fig.~\ref{fig:arch} illustrates the overall pipeline, transformer
block, and per-token attributes with role-based positional encoding.

\subsection{Learning Any-to-Any with Paired Data\label{subsec:training}}

\subsubsection{Training Objective}

We adopt the training objective of DR (Eq.~\ref{eq:paired_loss})
to learn all one-to-one mappings directly from paired datasets. For
any two modalities $i,j\in\mathcal{M}$ with a paired dataset $\Data_{i,j}$,
Modaliff learns both denoising directions, $i\to j$ and $j\to i$,
through a symmetric objective:
\begin{align}
\Loss_{\text{pair}}\left(\theta\right) & =\Expect_{\left(x^{i},x^{j}\right),t,\epsilon,\zeta}\left[\zeta\left\Vert z_{\theta}\left(z^{i}_{t},t,z^{j}\right)-z^{i}\right\Vert ^{2}_{2}\right.\nonumber \\
 & \ \ \ \ \ \ +\left.\left(1-\zeta\right)\left\Vert z_{\theta}\left(z^{j}_{t},t,z^{i}\right)-z^{j}\right\Vert ^{2}_{2}\right],\label{eq:paired_loss_z}
\end{align}

\noindent where $\epsilon\sim\Normal\left(0,\mathrm{I}\right)$, $t\sim\mathcal{U}\left(1,T\right)$,
$\zeta\sim\mathcal{B}(0.5)$, and $z^{i}_{t}=a_{t}z^{i}+\sigma_{t}\epsilon$
. Here, $z_{\theta}\left(z^{i}_{t},t,z^{j}\right)$ abbreviates Modaliff\textquoteright s
clean-latent prediction, computed on the unified sequence $\mathcal{H}\left(z^{j},z^{i}_{t}\right)$
of Eq.~\ref{eq:Ht}. The output tokens of the target segment, denoted
$\hat{h}^{i}_{\theta}\in\mathbb{R}^{N_{i}\times d}$ , are projected
back to the native latent space as:

\begin{equation}
z_{\theta}\left(z^{i}_{t},t,z^{j}\right)=\hat{h}^{i}_{\theta}\cdot W^{i}_{\text{out}},\ \ \ W^{i}_{\text{out}}\in\mathbb{R}^{d\times d_{i}}
\end{equation}
where $W^{i}_{\text{out}}$ is a trainable output projection matrix.
Unlike the noise-prediction in Eq.~\ref{eq:paired_loss}, Modaliff
predicts the clean latents, which is more robust across heterogeneous
pretrained latents with varying scales and heavy-tailed dimensions
that can destabilize noise prediction, as shown in Appdx.~\ref{subsec:Impact-of-parameterization}.
During training, we replace the condition with a null token $\varnothing$
with probability $0.2$, enabling CFG~\cite{ho2021classifier}.

To generate $x^{i}$ from $x^{j}$, we encode $z^{j}=\text{Enc}^{j}\left(x^{j}\right)$,
initialize $z^{i}_{T}\sim\mathcal{N}\left(0,\mathrm{I}\right)\in\mathbb{R}^{N_{i}\times d_{i}}$,
assemble $\mathcal{H}\left(z^{j},z^{i}_{T}\right)$, and run diffusion
sampler~\cite{song2020denoising} with CFG (Eq.~\ref{eq:cfg}) in
the noise space with $\epsilon_{\theta}\left(z^{i}_{t},t,\cdot\right)=\left(z^{i}_{t}-a_{t}z_{\theta}\left(z^{i}_{t},t,\cdot\right)\right)/\sigma_{t}$.
The final prediction $\hat{z}^{i}$ is decoded with the frozen decoder,
i.e., $x^{i}=\text{Dec}^{i}\left(\hat{z}^{i}\right)$.

\subsubsection{Learning with Missing Pairwise Data\label{subsec:Learning-with-Missing}}

In practice, some modality pairs are difficult to collect and provide
far less supervision than others. For example, image-audio dataset
($\Data_{i,j}$) \cite{chen2020vggsound} are typically scarce compared
with image-text ($\Data_{i,c}$) \cite{schuhmann2022laion} and audio-text
($\Data_{j,c}$) datasets \cite{gemmeke2017audio}, in which case
Modaliff can exploit the shared text modality $c$ to learn the missing
$X^{j}\leftrightarrow X^{i}$ mappings. To learn $X^{j}\to X^{i}$
(or similarly $X^{j}\leftarrow X^{i}$) without access to $\Data_{i,j}$,
we reformulate Eq.~\ref{eq:unpaired_loss} under clean-latent prediction
as:
\begin{align}
\mathcal{L}^{^{j\rightarrow i}}_{\text{unpair}}\left(\theta\right) & =\mathbb{E}_{\left(x^{j},x^{c}\right),t,z^{i}_{t}\sim p_{\theta}\left(z^{i}_{t}|z^{c}\right)}\left[\left\Vert z_{\theta}\left(z^{i}_{t},t,z^{j}\right)\right.\right.\nonumber \\
 & \ \ \ \ \ \ \ \ \ \ \ \ \ \ \ \ \ \left.\left.-\text{sg}\left(z_{\theta}\left(z^{i}_{t},t,z^{c}\right)\right)\right\Vert ^{2}_{2}\right],\label{eq:unpaired_loss_z}
\end{align}
where $\text{sg}\left(\cdot\right)$ denotes stop gradient. We jointly
optimize Eq.~\ref{eq:paired_loss_z} and Eq.~\ref{eq:unpaired_loss_z}
via Eq.~\ref{eq:final_loss}, enabling $X^{j}\to X^{i}$ without
$\Data_{i,j}$.

\subsubsection{Multi-Condition Generation via Score Composition\label{subsec:score_compose}}

Modaliff enables multi-condition generation by composing the single-condition
predictions learned from pairwise training. Let $i$ denote the target
modality and $\mathcal{S}\subseteq\left(\mathcal{M}\setminus i\right)$
the conditioning modalities. Assuming the conditions are conditionally
independent given the noisy target, i.e., $p\left(\left\{ z^{j}\right\} _{j\in\mathcal{S}}\mid z^{i}_{t}\right)=\prod_{j\in\mathcal{\mathcal{S}}}p(z^{j}\mid z^{i}_{t})$,
the product-of-experts formulation \cite{nguyen2025h} yields multi-condition
noise prediction as:
\begin{multline}
\tilde{\epsilon}_{\theta}\left(z^{i}_{t},t,\left\{ z^{j}\right\} _{j\in\mathcal{S}}\right)=\epsilon_{\theta}\left(z^{i}_{t},t,\varnothing\right)+\\
\sum_{j\in\mathcal{S}}w_{j}\left[\epsilon_{\theta}\left(z^{i}_{t},t,z^{j}\right)-\epsilon_{\theta}\left(z^{i}_{t},t,\varnothing\right)\right],\label{eq:composed_score}
\end{multline}
where $\epsilon_{\theta}\left(z^{i}_{t},t,\cdot\right)=\left(z^{i}_{t}-a_{t}z_{\theta}\left(z^{i}_{t},t,\cdot\right)\right)/\sigma_{t}$
and $\left\{ w_{j}\right\} _{j\in\mathcal{S}}$ control the contribution
of each condition. Eq.~\ref{eq:composed_score} extends single-condition
predictions to multi-condition generation without training on fully
aligned multimodal datasets and reduces to standard CFG when only
one condition is provided.

\subsection{Improving Multi-Condition Generation via Caption-Based Finetuning\label{subsec:multicond_finetune}}

Score composition (Eq.~\ref{eq:composed_score}) enables inference-time
multi-condition generation but limits the ability to capture interactions
among conditions and struggles with tasks such as $\left(X^{i}+X^{j}\right)\to X^{i}$,
where the required same-modality prediction $\epsilon_{\theta}\left(z^{i}_{t},t,z^{i}\right)$
is unavailable. To mitigate the gap, we propose a novel approach that
constructs a compact finetuning set for multi-condition generation
and uses it to adapt Modaliff. Our key idea is to encode joint semantics
of multiple conditions in language and transfer them through the text-to-target
mappings learned from one-to-one training. For example, image generation
from image and audio conditions can be reframed as text-to-image generation
using a caption that combines semantics of both conditions.

Formally, we consider two-condition generation $\left(X^{i}+X^{j}\right)\rightarrow X^{k}$,
with $i,j,k\in\mathcal{M}$, following prior any-to-any settings~\cite{tang2023any,li2025omniflow}.
Given two pairs $\left(x^{i},x^{c}\right)\sim\Data_{i,c}$ and $\left(x^{j},x'^{c}\right)\sim\Data_{j,c}$,
where $c$ denotes text, an instruction-tuned language model~\cite{yang2025qwen3}
merges $x^{c}$ and $x'^{c}$ into a natural joint description $\tilde{x}^{c}$
that preserves information from both $x^{i}$ and $x^{j}$, yielding
a finetuning triplet $\left(x^{i},x^{j},\tilde{x}^{c}\right)$. Random
pairing, however, provides a poor finetuning signal because unrelated
captions yield unnatural descriptions, while near-duplicate captions
contribute limited complementary information. We therefore compute
their cosine similarity as:
\begin{equation}
\text{sim}\left(x^{c},x'^{c}\right)=\frac{\left\langle \text{Emb}\left(x^{c}\right),\text{Emb}\left(x'^{c}\right)\right\rangle }{\left\Vert \text{Emb}\left(x^{c}\right)\right\Vert _{2}\left\Vert \text{Emb}\left(x'^{c}\right)\right\Vert _{2}},\label{eq:sim}
\end{equation}

\noindent where $\text{Emb}\left(\cdot\right)$ denotes a sentence-embedding
model \cite{reimers-2019-sentence-bert}, and retain only pairs satisfying
$\text{sim}\left(x^{c},x'^{c}\right)\in\left[\gamma_{\text{min}},\gamma_{\text{max}}\right]$,
where $\gamma_{\text{min}}$ and $\gamma_{\text{max}}$ control the
similarity range. The resulting triplets supervise task $\left(X^{i}+X^{j}\right)\rightarrow X^{k}$
with the objective:
\begin{equation}
\resizebox{\columnwidth}{!}{\ensuremath{{\displaystyle \Loss^{\left(i+j\right)\rightarrow k}_{\text{multi}}\left(\theta\right)=\Expect_{\left(x^{i},x^{j},\tilde{x}^{c}\right),\bar{z}^{k},z^{k}_{t}}\left[\left\Vert z_{\theta}\left(z^{k}_{t},t,\left\{ z^{i},z^{j}\right\} \right)-\bar{z}^{k}\right\Vert ^{2}_{2}\right],}}}\label{eq:multi}
\end{equation}
\begin{equation}
\text{s.t.\ }\begin{cases}
k=c: & z^{k}_{t}=a_{t}\tilde{z}^{c}+\sigma_{t}\epsilon,\ \ \ \ \bar{z}^{k}=\tilde{z}^{c}\\
k\neq c: & z^{k}_{t}\sim p_{\theta}\left(z^{k}_{t}\mid\tilde{z}^{c}\right),\ \ \bar{z}^{k}=\text{sg}\left(z_{\theta}\left(z^{k}_{t},t,\tilde{z}^{c}\right)\right)
\end{cases}\label{eq:multi_target}
\end{equation}
where $\tilde{z}^{c}=\text{Enc}^{c}\left(\tilde{x}^{c}\right)$. To
preserve the pretrained one-to-one mappings and finetune efficiently,
we optimize Eq.~\ref{eq:multi} using LoRA~\cite{hu2022lora}.

%% file: exp.tex
\subsection{Experiment Setup}

\subsubsection{Training Datasets}

For image-text dataset, we train Modaliff on 800K pairs sampled from
LAION-Aesthetics +6.25~\cite{schuhmann2022laion}, replacing noisy
captions containing URLs or HTML fragments with captions generated
by LLaVA~\cite{liu2023visual}. For audio-text dataset, we use 92K
human-annotated pairs from AudioCaps v2~\cite{kim2019audiocaps}.
Although Modaliff does not strictly require image-audio pairs, we
include 184K image-audio pairs from VGGSound~\cite{chen2020vggsound}
for fair comparison with baselines. We additionally train a variant
without VGGSound to assess image$\leftrightarrow$audio generation
through text as the shared modality. Full dataset details are provided
in Appx.~\ref{subsec:dataset}.

\begin{table}[t]
\begin{centering}
\setlength{\tabcolsep}{1mm}\fontsize{9}{11}\selectfont%
\begin{tabular}{ccccccc}
\toprule 
\multicolumn{1}{c}{Task} & T$\to$I & A$\to$I & I$\to$T & A$\to$T & T$\to$A & I$\to$A\tabularnewline
\midrule
\midrule 
Method & \multicolumn{2}{c}{FID$\downarrow$} & \multicolumn{2}{c}{CIDEr$\uparrow$} & \multicolumn{2}{c}{FAD$\downarrow$}\tabularnewline
\midrule 
SD3-Medium & 25.40 & - & - & - & - & -\tabularnewline
\midrule 
Sound2Vision & - & 42.55 & - & - & - & -\tabularnewline
\midrule 
LLaVA-Next & - & - & 109.3 & - & - & -\tabularnewline
\midrule 
Qwen2-Audio & - & - & - & 4.64 & - & -\tabularnewline
\midrule 
AudioLDM-L & - & - & - & - & 1.96 & -\tabularnewline
\midrule 
Seeing\&Hearing & - & - & - & - & - & 5.31\tabularnewline
\midrule 
CoDi{*} & \textcolor{black}{24.68} & 56.64 & 14.01 & 6.61 & 9.72 & 16.34\tabularnewline
\midrule 
OmniFlow{*} & \textcolor{black}{23.84} & 72.31 & \textcolor{black}{43.60} & 31.79 & \textcolor{black}{2.48} & 3.69\tabularnewline
\midrule 
FlowBind{*} & 17.39 & \textit{36.48} & 37.35 & \textcolor{black}{36.05} & 5.19 & \textit{2.52}\tabularnewline
\midrule 
Modaliff & \textbf{13.77} & \textbf{19.02} & \textit{44.47} & \textit{41.30} & \textbf{1.31} & \textbf{1.89}\tabularnewline
\midrule 
Modaliff-noVGG & \textit{14.18} & \textcolor{black}{57.23} & \textbf{46.69} & \textbf{41.57} & \textit{1.73} & \textcolor{black}{3.16}\tabularnewline
\bottomrule
\end{tabular}
\par\end{centering}
\caption{Fidelity metrics for cross-modal generation. {*} indicates results
that we reproduced using the official repository.\label{tab:main_result1}}
\end{table}

\subsubsection{Implementation Details}

We use CLIP-ViT-L16~\cite{radford2021learning} to encode an image
into $z^{\text{img}}\in\mathbb{R}^{1\times768}$ and Stable-UnCLIP-small~\cite{ramesh2022hierarchical}
to decode $z^{\text{img}}$. For text, we build an encoder based on
Qwen3-0.6B Embedding~\cite{zhang2025qwen3} to map each text sequence
to a fixed-length latent $z^{\text{txt}}\in\mathbb{R}^{32\times256}$,
and use a fine-tuned Qwen3-1.7B~\cite{yang2025qwen3} to generate
text from $z^{\text{txt}}$. Appx.~\ref{subsec:Text-Auto-Encoder}
provides details of the text encoder and text decoder. For audio,
we use CLAP-HTSAT~\cite{wu2023large} to encode each 10-second waveform
into $z^{\text{aud}}\in\mathbb{R}^{1\times512}$ and AudioLDM-S~\cite{liu2023audioldm}
to decode $z^{\text{aud}}$. Modality-specific encoders and decoders
are frozen during training. 

Following DiT-L~\cite{peebles2023scalable}, Modaliff uses 24 transformer
blocks, hidden dimension $d=1024$, and 16 attention heads, yielding
558M trainable parameters to fit a H100 80GB GPU. We train Modaliff
with Eq.~\ref{eq:paired_loss_z} for up to 500K iterations with a
batch size of 512, bfloat16 precision, and a constant learning rate
of 3e-5 with 5K warmup steps. With no VGGSound setting, we finetune
Modaliff following Sec.~\ref{subsec:Learning-with-Missing} to enable
image$\leftrightarrow$audio generation, with reduced learning rate
of 5e-6, and $\lambda_{1}=\lambda_{2}=1$ in Eq.~\ref{eq:final_loss},
as recommended by DR~\cite{kieu2026universal}. For multi-condition
generation, we construct 50K composed examples with $\gamma_{\text{min}}=0.4$,
$\gamma_{\text{max}}=0.8$, and finetune Modaliff with LoRA of rank
16 and scaling 32 for 60K iterations with a learning rate of 5e-6.
\begin{table}[t]
\begin{centering}
\setlength{\tabcolsep}{1mm}\fontsize{9}{11}\selectfont%
\begin{tabular}{ccccccc}
\toprule 
\multicolumn{1}{c}{Task} & T$\to$I & I$\to$T & A$\to$T & T$\to$A & A$\to$I & \multicolumn{1}{c}{I$\to$A}\tabularnewline
\midrule
\midrule 
Method & \multicolumn{2}{c}{CLIP$\uparrow$} & \multicolumn{2}{c}{CLAP$\uparrow$} & \multicolumn{2}{c}{ImgBind$\uparrow$}\tabularnewline
\midrule 
SD3-Medium & 31.60 & - & - & - & - & -\tabularnewline
\midrule 
Sound2Vision & - & - & - & - & 26.71 & -\tabularnewline
\midrule 
LLaVA-Next & - & 32.14 & - & - & - & -\tabularnewline
\midrule 
Qwen2-Audio & - & - & 17.09 & - & - & -\tabularnewline
\midrule 
AudioLDM-L & - & - & - & 36.05 & - & -\tabularnewline
\midrule 
Seeing\&Hearing & - & - & - & - & - & 30.69\tabularnewline
\midrule 
CoDi{*} & 30.32 & 26.1 & 17.86 & 10.23 & 15.96 & 5.53\tabularnewline
\midrule 
OmniFlow{*} & \textbf{31.70} & 27.8 & \textbf{48.65} & 33.13 & 8.56 & 13.91\tabularnewline
\midrule 
FlowBind{*} & 27.69 & \textit{28.38} & \textit{41.31} & 42.30 & \textit{24.43} & \textit{21.11}\tabularnewline
\midrule 
Modaliff & \textit{30.65} & 28.11 & 40.01 & \textbf{45.71} & \textbf{24.83} & \textbf{27.99}\tabularnewline
\midrule 
Modaliff-noVGG & 30.37 & \textbf{28.45} & 39.42 & \textit{44.01} & \textcolor{black}{12.83} & \textcolor{black}{13.79}\tabularnewline
\bottomrule
\end{tabular}
\par\end{centering}
\caption{Alignment scores for cross-modal generation. {*} indicates results
that we reproduced using the official repository.\label{tab:main_result2}}
\end{table}

\subsubsection{Evaluation settings}

We evaluate cross-modal generation across image (I), text (T), and
audio (A), using metrics for both fidelity and semantic alignment.
Image fidelity is measured by FID~\cite{heusel2017gans} on 256$\times$256
images, using 30K MSCOCO validation samples~\cite{lin2014microsoft}
for T$\to$I and VGGSound test samples for A$\to$I. Text quality
is evaluated with CIDEr~\cite{vedantam2015cider} on MSCOCO for I$\to$T,
where each image has five reference captions, and AudioCaps for A$\to$T,
with five captions per audio clip. Audio fidelity is measured by FAD~\cite{kilgour2019frechet}
on AudioCaps for T$\to$A and VGGSound for I$\to$A. Semantic alignment
is assessed using CLIP score~\cite{radford2021learning} for image-text
pairs, CLAP score~\cite{wu2023large} for audio-text pairs, and ImgBind
score~\cite{girdhar2023imagebind} for audio-image pairs. For multi-condition
evaluation, we hold out 1K samples from the finetuning set, using
$\gamma_{\text{min}}=0.4$, $\gamma_{\text{max}}=0.8$ to encourage
semantically distinct conditions.

\subsubsection{Baselines}

We compare Modaliff against state-of-the-art specialist models and
any-to-any diffusion models, including CoDi~\cite{tang2023any},
OmniFlow~\cite{li2025omniflow}, and FlowBind~\cite{cha2026flowbind}.
For specialists, we adopt the results from FlowBind. For any-to-any
models, we use the official checkpoints, sampling scripts, recommended
hyperparameters, and fixing latent generation to 50 sampling steps
across all methods for a fair comparison.

\begin{figure*}[t]
\begin{centering}
\includegraphics[width=1\textwidth]{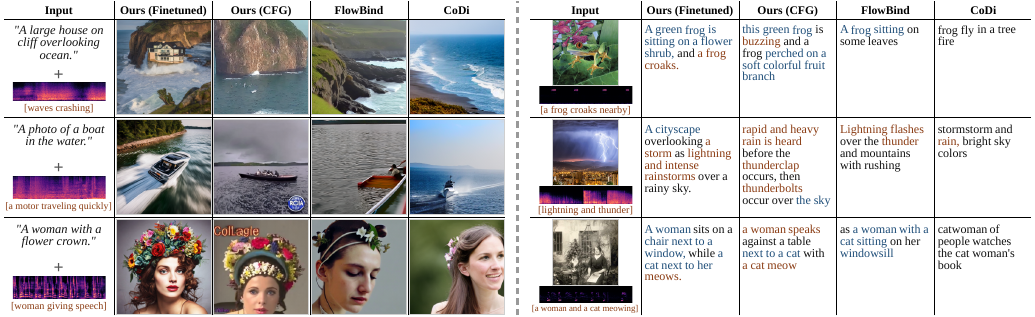}
\par\end{centering}
\caption{Qualitative results of Modaliff and generalists on (T + A)$\to$I
and (I + A)$\to$T.\label{fig:vis_multicond}}
\end{figure*}

\subsection{Experiment Results}

\subsubsection{One-to-One Cross-Modality Generation}

Tabs. \ref{tab:main_result1} and \ref{tab:main_result2} report fidelity
and semantic alignment for one-to-one cross-modal generation across
text, image, and audio, while Appdx. \ref{subsec:dataset} compares
the training-data scale and trainable parameters of the generalists.
We observe that Modaliff outperforms the generalists in fidelity on
most generation tasks, with particularly large gains on T$\to$I and
T$\to$A, highlighting the effectiveness of preserving modality-native
token representations within a unified denoising sequence. CoDi\textquoteright s
inferior performance typically stems from combining multiple outdated
U-Net architectures and training under a multi-task objective, making
it difficult to balance learning across modalities. OmniFlow exhibits
strong T$\to$I alignment, which is expected given its initialization
from SD3-Medium, a strong foundation model trained on billions of
high-quality samples. However, OmniFlow underperforms in FID, which
we attribute to the substantial amount of synthetic and stylized imagery
in the training data of SD3-Medium, causing its generations to differ
in visual style from the realistic images in MSCOCO. Moreover, both
CoDi and OmniFlow denoise high-dimensional image and audio VAE latents~\cite{rombach2022high,liu2023audioldm},
requiring larger denoisers and higher training cost. In contrast,
Modaliff operates on compact semantic latents produced by contrastive
encoders (CLIP for images and CLAP for audio), shifting low-level
synthesis to frozen decoders and matching these baselines with a substantially
smaller denoiser and fewer computational resources. Although FlowBind
improves performance over CoDi by jointly training all velocity models
with a shared latent encoder, its fixed shared vector representation
discards the token-level structure of text, causing FlowBind to underperform
in both text generation quality and alignment on the text-conditioned
routes T$\to$I and T$\to$A.

When VGGSound is excluded from training, Modaliff predictably degrades
on A$\to$I and I$\to$A, as Modaliff must learn these mappings indirectly
through text rather than from direct image-audio supervision. Nevertheless,
Modaliff-noVGG still outperforms CoDi on I$\to$A and OmniFlow on
both A$\to$I and I$\to$A in fidelity, while remaining comparable
in alignment despite both baselines using millions of image-audio
pairs, indicating that Modaliff can effectively use text as the central
modality to learn image-audio generation.

\subsubsection{Multi-Condition Generation}

Tab.~\ref{tab:multi_cond_results} reports multi-condition alignment
for generalists and Modaliff using score composition (Sec.~\ref{subsec:score_compose})
and composed caption finetuning (Sec.~\ref{subsec:multicond_finetune}).
For each task, we measure the ImageBind score between the generated
output and each condition separately. Without multi-condition training,
Modaliff remains competitive with the baselines and performs particularly
well on (T+I)$\to$A, demonstrating that the learned one-to-one mappings
can be composed effectively at inference. Finetuning with composed
captions further improves five of the six alignment scores and achieves
the best performance on four, confirming that language provides effective
supervision for modeling interactions between conditions. The only
exception is the lower A-I score of finetuned variant on (T+A)$\to$I
relative to Modaliff (CFG), which arises because static images are
inherently limited in depicting transient acoustic events such as
speaking or chirping, even when this information is preserved in the
composed caption, thereby limiting their transfer from the T$\to$I
teacher to (T+A)$\to$I mapping. 

\begin{figure}[t]
\begin{centering}
\includegraphics[width=0.465\textwidth]{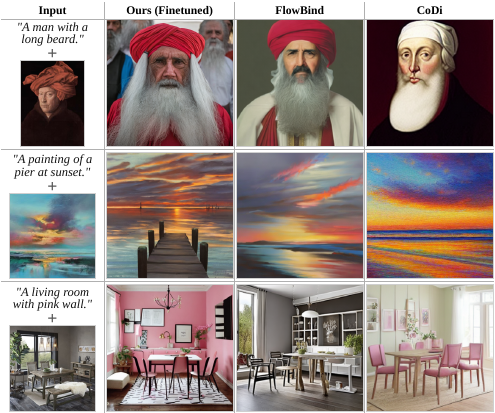}
\par\end{centering}
\caption{Qualitative results of Modaliff and generalists on (T + I)$\to$I.\label{fig:vis_ti2i}}
\end{figure}

This behavior is consistent with Fig.~\ref{fig:vis_multicond}, where
fine-tuning improves visual quality and text alignment, while actions
suggested by the audio, such as giving a speech, remain difficult
to depict in a single image. For (T + I)$\to$I, Fig.~\ref{fig:vis_ti2i}
shows that Modaliff produces higher-quality images and follows both
conditions more accurately, demonstrating the effectiveness of the
proposed finetuning strategy.

\begin{table}[t]
\begin{centering}
\setlength{\tabcolsep}{1mm}\fontsize{9}{11}\selectfont%
\begin{tabular}{ccccccc}
\toprule 
\multicolumn{1}{c}{} & \multicolumn{2}{c}{(T+A)$\to$I} & \multicolumn{2}{c}{(I+A)$\to$T} & \multicolumn{2}{c}{(T+I)$\to$A}\tabularnewline
\midrule 
Method & T-I & A-I & I-T & A-T & T-A & I-A\tabularnewline
\midrule
\midrule 
CoDi & 22.97 & \textbf{22.93} & 15.05 & \textit{19.51} & 8.21 & 3.30\tabularnewline
\midrule 
FlowBind & \textit{27.71} & 20.82 & 20.21 & \textbf{20.54} & 12.92 & 9.60\tabularnewline
\midrule 
Modaliff (CFG) & 23.39 & \textit{20.93} & \textit{21.92} & 17.13 & \textit{20.89} & \textit{13.89}\tabularnewline
\midrule 
Modaliff (Finetuned) & \textbf{29.41} & 16.36 & \textbf{24.28} & 18.35 & \textbf{22.10} & \textbf{14.08}\tabularnewline
\bottomrule
\end{tabular}
\par\end{centering}
\caption{ImgBind scores for multi-condition generation. This capability is
unavailable in OmniFlow\textquoteright s released code.\label{tab:multi_cond_results}}
\end{table}

\subsection{Ablation Studies}

\subsubsection{Impact of Role-based Positional Embedding}

\begin{figure}
\subfloat[CIDEr on I$\to$T and A$\to$T]{\begin{centering}
\includegraphics[width=0.22\textwidth]{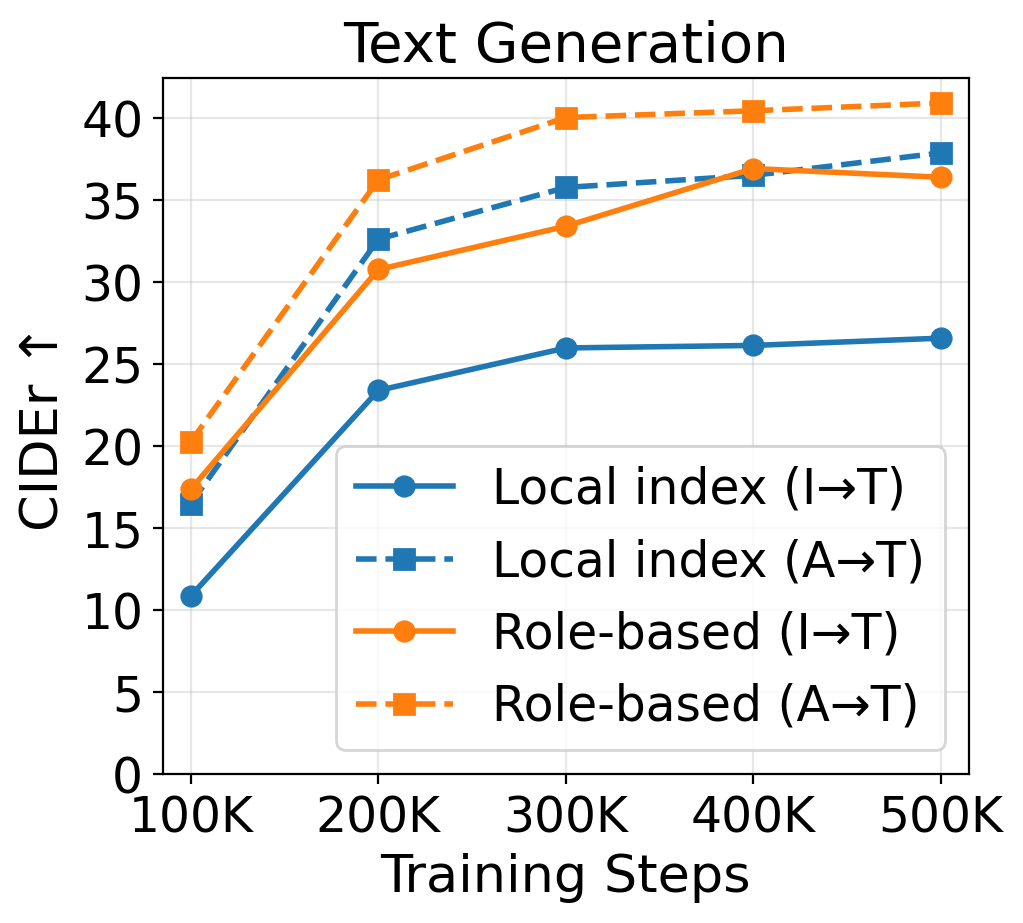}
\par\end{centering}

}\subfloat[FID 5K on T$\to$I]{\begin{centering}
\includegraphics[width=0.22\textwidth]{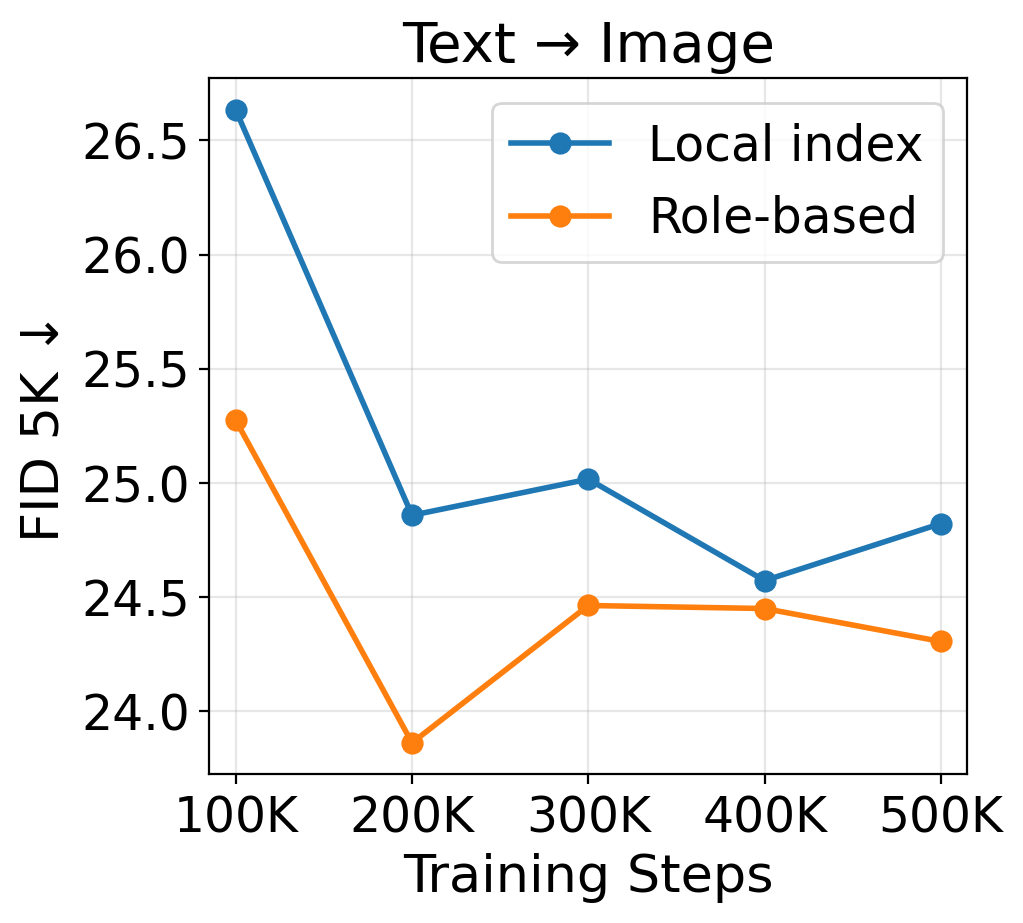}
\par\end{centering}
}\caption{Learning curves comparing role-based positional encoding and local
indexing.\label{fig:learning_curves_PE}}
\end{figure}

We compare role-based PE against local indexing using two Modaliff
models that differ only in the positional scheme, each reduced to
159M parameters for efficiency. As shown in Figs.~\ref{fig:learning_curves_PE},
role-based PE consistently outperforms local indexing on all generation
routes, with the largest gain on text generation. We attribute this
improvement to the positional conflict introduced by local indexing
in the attention projections shared across all generation tasks. Under
local indexing, each condition and target segment resets its positions
from the same origin, creating artificial correspondences between
unrelated tokens that share similar indices. The shared projections
must then suppress this bias while preserving the positional structure
of the target sequence, which is particularly important for text generation.
Role-based PE removes these cross-segment biases by presenting condition
tokens to target queries under a shared positional reference, enabling
more effective routing across variable generation tasks.

\subsubsection{Impact of the Similarity Band $\left[\gamma_{\text{min}},\gamma_{\text{max}}\right]$ }

To evaluate similarity filtering for multi-condition triplet construction,
we fix $\gamma_{\text{max}}=0.8$ to exclude near-duplicate pairs,
vary $\gamma_{\text{min}}\in\left\{ 0,0.2,0.4,0.6\right\} $, and
finetune Modaliff on 10K composed samples per setting. Tab.~\ref{tab:vary_gamma_min}
reports the alignment between the generated target and each of the
two conditions across three multi-condition tasks. Across all tasks,
increasing $\gamma_{\text{min}}$ improves alignment with one condition
but reduces alignment with the other. This is because a high $\gamma_{\text{min}}$
retains pairs with highly similar captions, causing $\tilde{x}^{c}$
to emphasize shared content and provide limited supervision for composing
distinct conditions. Conversely, a lower $\gamma_{\text{min}}$ admits
more diverse pairs and strengthens the use of complementary information
but may introduce details supported by only one input, thereby reducing
alignment with the other. These results support our hypothesis for
similarity-based filtering and we use $\gamma_{\text{min}}=0.4$ to
balance individual-condition alignment with effective multi-condition
fusion.

\begin{table}
\begin{centering}
\setlength{\tabcolsep}{1mm}\fontsize{9}{11}\selectfont%
\begin{tabular}{ccccccc}
\toprule 
\multicolumn{1}{c}{} & \multicolumn{2}{c}{(T+A)$\to$I} & \multicolumn{2}{c}{(I+A)$\to$T} & \multicolumn{2}{c}{(T+I)$\to$A}\tabularnewline
\midrule 
$\gamma_{\text{min}}$ & T-I & A-I & I-T & A-T & T-A & I-A\tabularnewline
\midrule
\midrule 
0 & 26.80 & 16.42 & 23.23 & 14.98 & 16.28 & 12.71\tabularnewline
\midrule 
0.2 & 27.13 & 16.36 & 23.55 & 14.35 & 16.43 & 12.33\tabularnewline
\midrule 
0.4 & 27.25 & 15.60 & 23.90 & 13.65 & 16.84 & 11.78\tabularnewline
\midrule 
0.6 & 27.78 & 15.37 & 23.98 & 12.13 & 17.04 & 11.20\tabularnewline
\bottomrule
\end{tabular}
\par\end{centering}
\caption{ImgBind score between the target and each conditions with $\gamma_{\text{min}}\in\left\{ 0,0.2,0.4,0.6\right\} $
and $\gamma_{\text{max}}=0.8$. Each setting finetunes on 10K samples.\label{tab:vary_gamma_min}}
\end{table}

%% file: relate.tex
Building on the success in pairwise cross-modal generation \cite{liu2023audioldm,esser2024scaling,le2026framedit},
diffusion models~\cite{song2020score} have recently been extended
to unified multimodal generation. Versatile Diffusion~\cite{xu2023versatile}
extends latent diffusion into a unified image-text framework with
a modular cross-modal architecture, but its model size grows with
the number of modalities and the framework remains limited to the
image-text setting. UniDiffuser~\cite{bao2023one} instead models
the joint distribution of all modalities using a single transformer-based
noise prediction network that represents each modality as a token
sequence. This design provides a general formulation for multimodality,
but it requires fully aligned multimodal tuples and substantial training
time. CoDi~\cite{tang2023any} extends any-to-any generation to diverse
modalities, including text, image, audio, and video, through a two-stage
pipeline that first learns modality-specific encoders with contrastive
learning and then trains diffusion models for conditional generation.
While this supports flexible and multi-source generation, it requires
contrastive pretraining and multiple diffusion models whose number
scales linearly with the number of modalities. OmniFlow~\cite{li2025omniflow}
replaces diffusion with flow matching to model a joint distribution
across modalities and adapts a pretrained SD3 backbone~\cite{esser2024scaling}
to reduce training cost. However, it still depends on fully aligned
multimodal tuples, and missing modalities must be synthesized during
training. More recently, FlowBind~\cite{cha2026flowbind} adopts
flow matching to learn a shared latent prior and modality-specific
velocity models that map between the prior and each modality\textquoteright s
latent space. Although FlowBind jointly optimizes all models from
paired data, it requires matched dimensions between the prior and
modality representations, scales one model per modality, and assumes
paired supervision for every modality pair.

In summary, most of existing multimodal diffusion models rely on fully
aligned tuples, contrastive pretraining, or modality-specific generators
whose parameters scale linearly with the number of modalities. FlowBind
reduces reliance on aligned tuples through single-stage paired-data
training, but couples all modalities to a shared prior's dimensionality,
uses one velocity model per modality, and assumes supervision for
every modality pair. By contrast, our method learns directly from
paired data, preserves modality-specific structure in native latent
tokens, uses a single denoiser for all generation tasks to avoid modality-dependent
parameter growth, and supports settings with missing modality pairs.

%% file: discuss.tex
We presented Modaliff, a token-based diffusion framework for any-to-any
generation that requires neither a shared representation space nor
fully aligned multimodal tuples. By representing modality-native condition
and target tokens under a unified token sequence, Modaliff supports
diverse one-to-one and multi-condition tasks using a single denoiser
trained on paired data. Extensive experiments across text, image,
and audio demonstrate that Modaliff achieves the best fidelity among
any-to-any models on most generation tasks while using substantially
fewer parameters and training data.

%% file: appdx.tex
\section{Additional Details of the Proposed Framework}

\subsection{Comparison between existing diffusion-based multi-modal generation
frameworks and ours\label{subsec:dataset}}

\begin{table}
\begin{centering}
\resizebox{\textwidth}{!}{%
\par\end{centering}
\begin{centering}
\begin{tabular}{|>{\centering}m{0.12\textwidth}|>{\centering}m{0.15\textwidth}|>{\centering}m{0.1\textwidth}|>{\centering}m{0.1\textwidth}|>{\centering}m{0.1\textwidth}|>{\centering}m{0.1\textwidth}|>{\centering}m{0.12\textwidth}|>{\centering}m{0.1\textwidth}|}
\hline 
\multirow{2}{0.12\textwidth}{Method} & \multicolumn{2}{c|}{Text} & \multicolumn{2}{c|}{Image} & \multicolumn{2}{c|}{Audio} & \multirow{2}{0.1\textwidth}{Diffusion Model}\tabularnewline
\cline{2-7}
 & Encoder & Decoder & Encoder & Decoder & Encoder & Decoder & \tabularnewline
\hline 
\hline 
Versatile Diffusion \cite{xu2023versatile} & \multicolumn{2}{c|}{Optimus \cite{li2020optimus} (227M)} & \multicolumn{2}{c|}{\begin{cellvarwidth}[m]
\centering
VQ-VAE \cite{rombach2022high} 

(84M)
\end{cellvarwidth}} & - & - & 2D+1D UNet

(2.6B)\tabularnewline
\hline 
UniDiffuser \cite{bao2023one} & CLIP-L \cite{radford2021learning}

(124M) & GPT2 \cite{radford2019language} (124M) & \multicolumn{2}{c|}{\begin{cellvarwidth}[m]
\centering
VQ-VAE \cite{rombach2022high} 

(84M)
\end{cellvarwidth}} & - & - & U-ViT \cite{bao2023all}

(952M)\tabularnewline
\hline 
CoDi \cite{tang2023any} & \multicolumn{2}{c|}{Optimus \cite{li2020optimus} (227M)} & \multicolumn{2}{c|}{\begin{cellvarwidth}[m]
\centering
VQ-VAE \cite{rombach2022high} 

(84M)
\end{cellvarwidth}} & \multicolumn{2}{c|}{\begin{cellvarwidth}[m]
\centering
AudioLDM-VAE \cite{liu2023audioldm}

(111M)
\end{cellvarwidth}} & 2D+1D UNet

(4.3B)\tabularnewline
\hline 
OmniFlow \cite{li2025omniflow} & CLIP-L + CLIP-G + 

Flan-T5-L \cite{li2025omniflow} (783M) & TinyLlama \cite{zhang2024tinyllama} (1.1B) & \multicolumn{2}{c|}{\begin{cellvarwidth}[m]
\centering
VQ-VAE \cite{rombach2022high} 

(84M)
\end{cellvarwidth}} & \multicolumn{2}{c|}{\begin{cellvarwidth}[m]
\centering
AudioLDM-VAE \cite{liu2023audioldm} 

(111M)
\end{cellvarwidth}} & SD3-DiT

(3.4B)\tabularnewline
\hline 
FlowBind \cite{cha2026flowbind} & Embedding Gemma \cite{vera2025embeddinggemma} 

(300M) & Gemma3 \cite{gemma2025gemma3}

(1B) & CLIP-L Image Enc \cite{radford2021learning} (304M) & Stable-unCLIP-small \cite{ramesh2022hierarchical} (1.6B) & CLAP-HTSAT + Autoencoder \cite{wu2023large}

(200M) & AudioLDM-M \cite{liu2023audioldm}

(652M) & MLP

(568M)\tabularnewline
\hline 
Modaliff (ours) & Qwen3-Embedding \cite{zhang2025qwen3}

(600M) & Qwen3 \cite{yang2025qwen3} 

(1.7B) & CLIP-L Image Enc \cite{radford2021learning} (304M) & Stable-unCLIP-small \cite{ramesh2022hierarchical} (1.6B) & CLAP-HTSAT \cite{wu2023large}

(158M) & AudioLDM-S \cite{liu2023audioldm}

(421M) & Multimodal Diffusion Transformer

(558M)\tabularnewline
\hline 
\end{tabular}}
\par\end{centering}
\caption{Comparison between the components for each modality used in our methods
and baselines.\label{tab:model_compare}}
\end{table}

\begin{table}
\begin{centering}
\resizebox{\textwidth}{!}{
\begin{tabular}{|c|c|c|c|c|c|c|c|}
\hline 
\multirow{2}{*}{Method} & \multirow{2}{*}{Data format} & \multicolumn{2}{c|}{Text-Image} & \multicolumn{2}{c|}{Text-Audio} & \multicolumn{2}{c|}{Image-Audio}\tabularnewline
\cline{3-8}
 &  & Name & Size & Name & Size & Name & Size\tabularnewline
\hline 
\hline 
\begin{cellvarwidth}[m]
\centering
Versatile 

Diffusion \cite{xu2023versatile}
\end{cellvarwidth} & Pairwise & \begin{cellvarwidth}[m]
\centering
LAION-2B-en

COYO-700M
\end{cellvarwidth} & 2.7B & - & - & - & -\tabularnewline
\hline 
UniDiffuser \cite{bao2023one} & Pairwise & \begin{cellvarwidth}[m]
\centering
LAION-2B-en

LAION-high-resolution

LAION-aesthetics v2 5+
\end{cellvarwidth} & 2.4B & - & - & - & -\tabularnewline
\hline 
CoDi \cite{tang2023any} & \begin{cellvarwidth}[m]
\centering
Aligned tuples

+ Pairwise
\end{cellvarwidth} & LAION-400M & 400M & \begin{cellvarwidth}[m]
\centering
AudioSet

AudioCaps

Freesound

BBCSoundEffect
\end{cellvarwidth} & 3.5M & \begin{cellvarwidth}[m]
\centering
AudioSet

SoundNet
\end{cellvarwidth} & 2M\tabularnewline
\hline 
OmniFlow \cite{li2025omniflow} & \begin{cellvarwidth}[m]
\centering
Aligned tuples

+ Pairwise
\end{cellvarwidth} & \begin{cellvarwidth}[m]
\centering
LAION-aesthetics-3M

CC12M

COYO-700M (Subset)

LAION-COCO

T2I-2M
\end{cellvarwidth} & 28M & \begin{cellvarwidth}[m]
\centering
AudioSet

AudioCaps

WavCaps
\end{cellvarwidth} & 2.4M & \begin{cellvarwidth}[m]
\centering
VGGSound

SoundNet
\end{cellvarwidth} & 2.2M\tabularnewline
\hline 
FlowBind \cite{cha2026flowbind} & Pairwise & \begin{cellvarwidth}[m]
\centering
LAION-COCO

Flickr-30K
\end{cellvarwidth} & 272K & AudioCaps v2 & 92K & VGGSound & 184K\tabularnewline
\hline 
Modaliff (ours) & Pairwise & LAION-aesthetics 6.25+ & 800K & AudioCaps v2 & 92K & \begin{cellvarwidth}[m]
\centering
VGGSound 

(Optional)
\end{cellvarwidth} & \begin{cellvarwidth}[m]
\centering
184K

(0K)
\end{cellvarwidth}\tabularnewline
\hline 
\end{tabular}}
\par\end{centering}
\caption{Comparison between the multi-modal training datasets used in our methods
and baselines.\label{tab:data_compare}}
\end{table}

In Tabs.~\ref{tab:model_compare} and \ref{tab:data_compare}, we
detail the framework designs and the training data used by representative
multimodal diffusion based methods, including Versatile Diffusion~\cite{xu2023versatile},
UniDiffuser~\cite{bao2023one}, CoDi~\cite{tang2023any}, OmniFlow~\cite{li2025omniflow},
and FlowBind~\cite{cha2026flowbind}. For each framework, we highlight
important architectural designs that determine how modalities are
processed and modeled, including the modality specific encoders and
decoders, and the backbone model for parameterizing the score function
of diffusion models. We also list the supervision format required
by each framework, such as fully aligned multimodal tuples or pairwise
datasets, together with the corresponding training dataset and the
number of training samples for each pairwise generation task.

From Tabs.~\ref{tab:model_compare} and~\ref{tab:data_compare},
Modaliff uses a 558M trainable denoiser and approximately 1.1M paired
samples for the full text-image-audio setting, requiring substantially
fewer trainable parameters and less data than large-scale any-to-any
models such as CoDi and OmniFlow. This comparison highlights the efficiency
of Modaliff in predicting compact semantic latents from frozen contrastive
pretrained encoders such as CLIP and CLAP while delegating low-level
synthesis to frozen modality-specific decoders. FlowBind provides
the closest comparison in trainable model scale, but requires a common
vector dimensionality across the shared prior and modality representations
and uses a separate MLP for each modality. Modaliff instead preserves
each modality\textquoteright s native latent token sequence and aligns
only the transformer width through lightweight projections, allowing
one shared denoiser to process heterogeneous token lengths and feature
dimensions. Among methods supporting image-audio generation, CoDi,
OmniFlow, and FlowBind rely on direct image-audio pairs, whereas Modaliff
can learn image-to-audio and audio-to-image mappings through text
as the shared modality when VGGSound is omitted.

\subsection{Details of the Proposed Text Auto-Encoder\label{subsec:Text-Auto-Encoder}}

\subsubsection{High-fidelity Text Autoencoder}

Since diffusion-based multimodal generation frameworks usually operate
in latent space, their performance depends on the quality of the underlying
auto-encoders for each modality. While effective auto-encoders exist
for images and audio (e.g.,~\cite{rombach2022high} and~\cite{liu2023audioldm}),
comparable solutions for text remain underdeveloped. We empirically
observe that current frameworks~\cite{tang2023any,bao2023one,li2025omniflow}
often produce repetitive and unnatural text. We attribute this to
the use of outdated or underpowered text models. For example, Versatile
Diffusion~\cite{xu2023versatile} adopts Optimus~\cite{li2020optimus},
which pairs a BERT encoder with a GPT-2 decoder, while OmniFlow~\cite{li2025omniflow}
employs a CLIP-L + CLIP-G + Flan-T5-L encoder with a TinyLlama-1.1B
decoder~\cite{zhang2024tinyllama}.

To address this limitation, we introduce a text auto-encoder that
uses Qwen3-0.6B Embedding~\cite{zhang2025qwen3} as the encoder and
Qwen3-1.7B~\cite{yang2025qwen3} as the decoder. As the latest generation
in the Qwen family, Qwen3 exhibits strong multilingual generation
and reasoning capabilities, while the embedding counterpart achieves
competitive performance on text representation and retrieval benchmarks.
Together, they form a high-fidelity text auto-encoding backbone.

Given an input sequence $x=\left(x_{1},...,x_{L}\right)$, we first
obtain token-level representations $h\in\Real^{L\times d_{\text{enc}}}$
from the Qwen3 Embedding model without applying any pooling operation.
Then, we employ a Flamingo-style resampler~\cite{alayrac2022flamingo}
to compress the variable-length sequence into a fixed-size latent
representation $z\in\mathbb{R}^{M\times d_{\text{latent}}}$. A lightweight
projector subsequently maps $z$ to a sequence of prefix embeddings
$r\in\mathbb{R}^{M\times d_{\text{dec}}}$, which are prepended to
the language model decoder as soft prompts for reconstruction. During
training, we freeze the embedding encoder while jointly optimizing
the resampler, projector, and decoder. To improve parameter efficiency,
we finetune the decoder using LoRA~\cite{hu2022lora} while keeping
the remaining decoder weights frozen. The model is trained with the
standard next-token prediction objective:
\begin{equation}
\mathcal{L}\left(\phi\right)=-\sum^{L}_{l=1}\log p_{\phi}\left(x_{l}\mid r,x_{<l}\right).\label{eq:ntp}
\end{equation}

\subsubsection{Training Configurations}

\begin{figure}[t]
\begin{centering}
\subfloat[Pipeline of text auto-encoder.]{\begin{centering}
\includegraphics[width=0.3\textwidth]{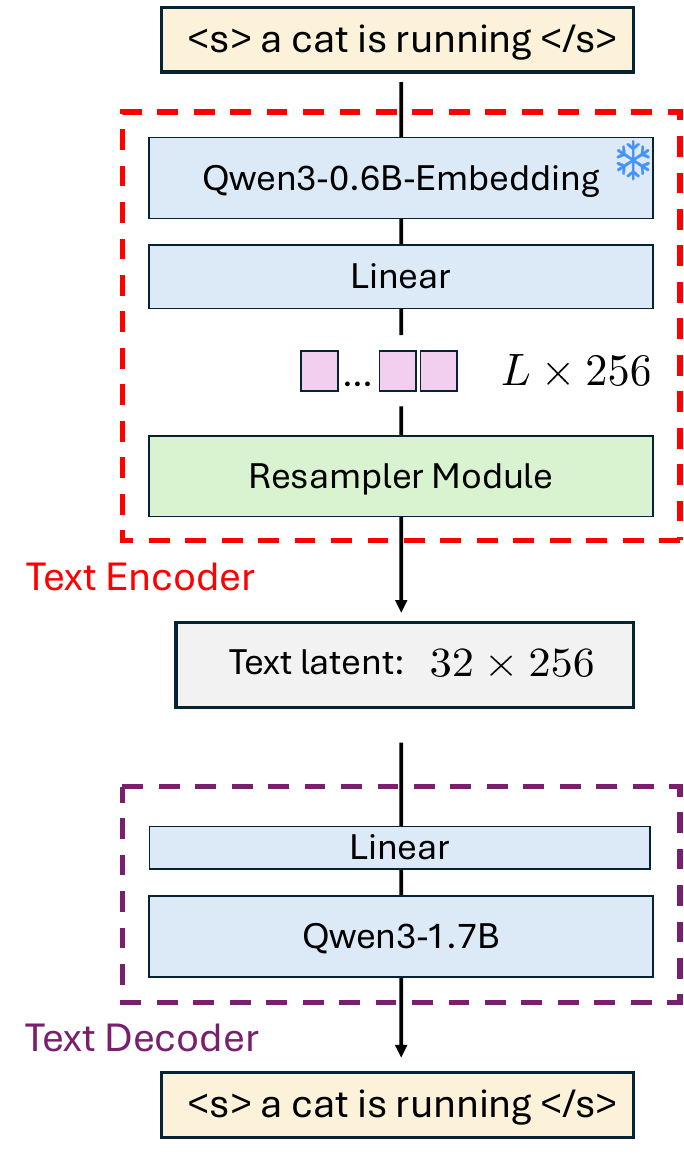}
\par\end{centering}
}\qquad{}\qquad{}\qquad{}\subfloat[Resampler module.]{\begin{centering}
\includegraphics[width=0.35\textwidth]{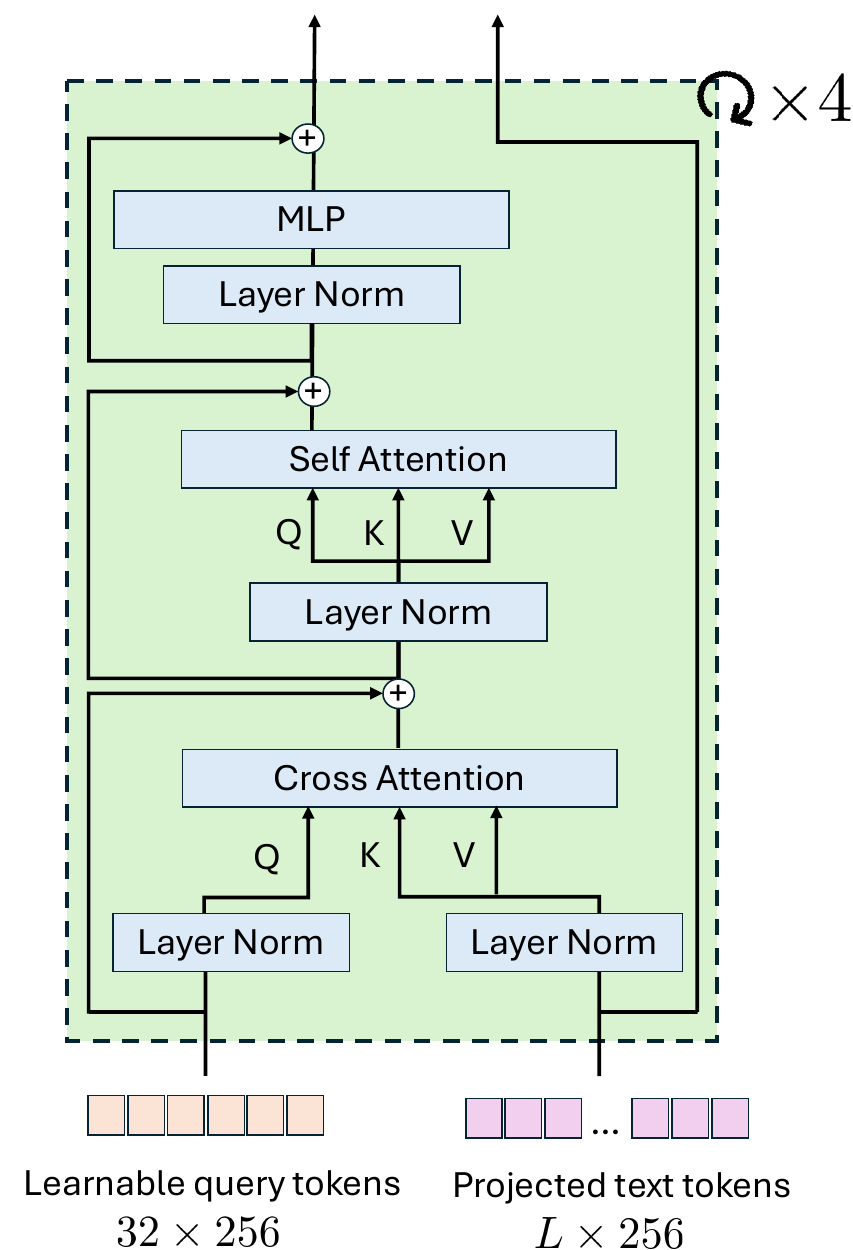}
\par\end{centering}
}
\par\end{centering}
\caption{\label{fig:text_ae}Architecture of Text Auto-Encoder: a) Overall
pipeline of the proposed text auto-encoder. b) Details of the resampler
module to produce latent embedding of text.}
\end{figure}

We use Qwen3-0.6B Embedding without a pooling layer to encode a text
sequence $x=\left(x_{1},...,x_{L}\right)$ into token-level representations
$h\in\Real^{L\times d_{\text{enc}}}$, where $L$ is the sequence
length and $d_{\text{enc}}=1024$ follows the Qwen3-0.6B Embedding
specification. We then apply a linear projection to obtain $h'\in\Real^{L\times d_{\text{latent}}}$,
reducing the feature dimension to $d_{\text{latent}}=256$ while retaining
sufficient information for accurate reconstruction. To convert variable-length
token sequences into a fixed-length latent, we adopt the resampler
module from Flamingo~\cite{alayrac2022flamingo} to compress $h'$
into $z$ containing $M$ learned query tokens, i.e., $z\in\mathbb{R}^{M\times d_{\text{latent}}}$,
with $M=32$ following common practice in BLIP-2~\cite{li2023blip2}
and OmniFlow~\cite{li2025omniflow}. These query tokens $z$ serve
as the text latent representation for downstream diffusion models.
For decoding, we map $z$ through a linear projector to obtain a $32\times2048$
prefix embedding that is fed into a Qwen3-1.7B decoder. The full architecture
is illustrated in Fig.~\ref{fig:text_ae}.

We train the text autoencoder using the objective in Eq.~\ref{eq:ntp}
with a two-stage procedure. In the first stage, we freeze the Qwen3-0.6B
Embedding encoder, train the resampler and projection layers, and
finetune the Qwen3-1.7B decoder with LoRA using rank $r=16$ and scaling
$\alpha=32$. We train for five epochs on captions from the text-image
dataset and exclude the text-audio captions, since they are often
short phrases rather than well-formed sentences. We use a learning
rate of 1e-5 and a global batch size of 64 with the AdamW optimizer~\cite{Loshchilov2019}.
In the second stage, we freeze the resampler and continue finetuning
the Qwen3-1.7B decoder with LoRA for three additional epochs on the
merged caption corpus of Tab.~\ref{tab:data_compare}.

\section{Further Experimental Results}

\subsection{Study on different text encoders}

\begin{figure}[b]
\subfloat[Perplexity]{\begin{centering}
\includegraphics[width=0.3\textwidth]{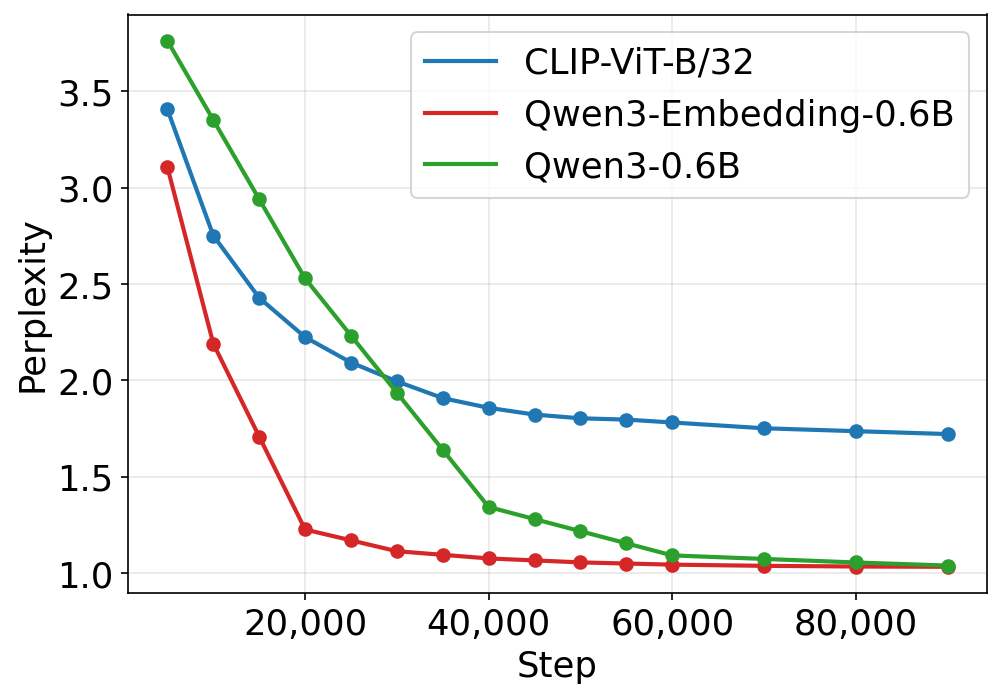}
\par\end{centering}
}\hfill{}\subfloat[BLEU-1]{\begin{centering}
\includegraphics[width=0.3\textwidth]{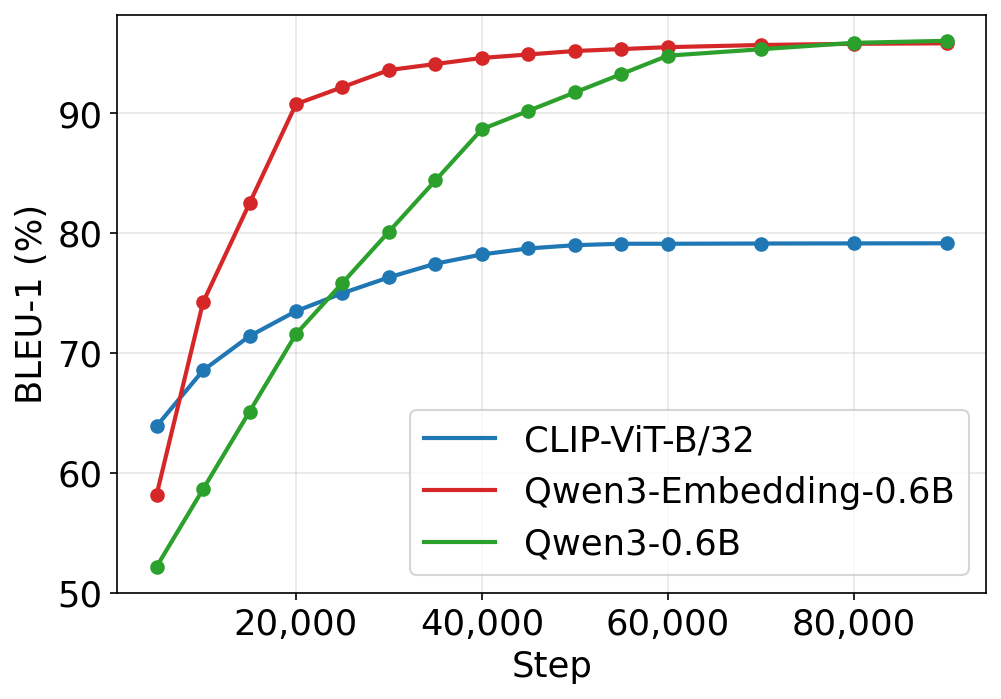}
\par\end{centering}
}\hfill{}\subfloat[BLEU-4]{\begin{centering}
\includegraphics[width=0.3\textwidth]{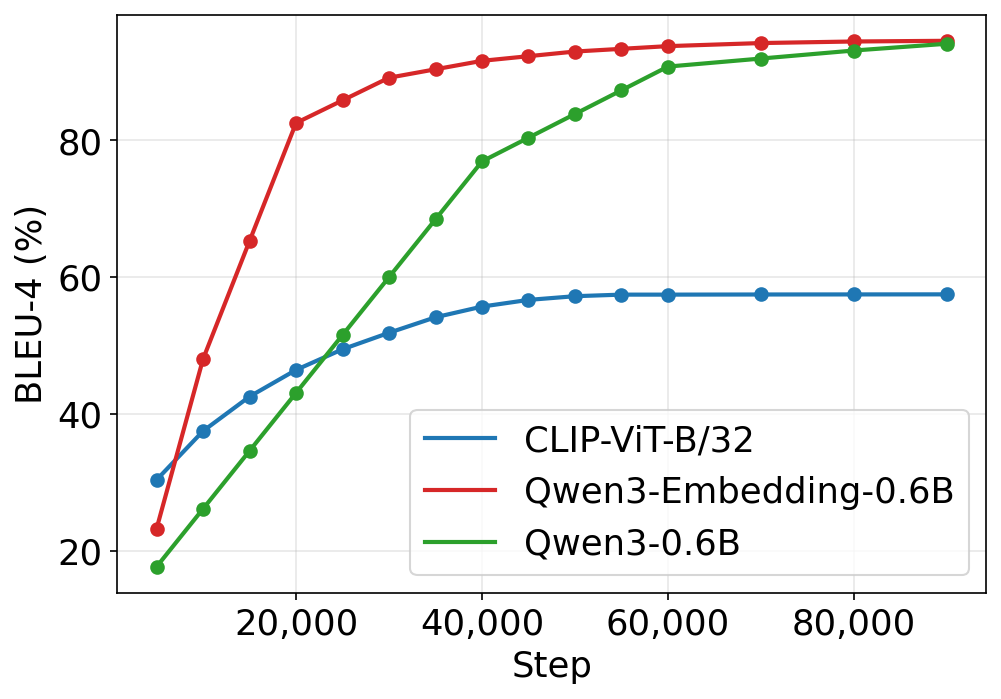}
\par\end{centering}
}

\caption{Validation reconstruction metrics under different text encoder choices
with the same resampler and Qwen3-1.7B decoder. We report Perplexity,
BLEU-1, and BLEU-4 over training for Qwen3-Embedding-0.6B, Qwen3-0.6B,
and CLIP-ViT-B/32.\label{fig:reconstruct_text}}
\end{figure}

We examine how the choice of text encoder affects the performance
of our text auto-encoder model by training the same architecture with
three encoder options, Qwen3-Embedding-0.6B, Qwen3-0.6B, and CLIP-ViT-B/32,
while keeping the resampler design and the Qwen3-1.7B text decoder
fixed. Fig.~\ref{fig:reconstruct_text} reports reconstruction metrics
on the validation set throughout training, including Perplexity, BLEU-1,
and BLEU-4. Overall, the Qwen3 based encoders consistently achieve
lower perplexity and higher BLEU scores than CLIP, indicating stronger
reconstruction quality. This is expected because the decoder is a
Qwen3 model, so encoders from the same model family produce representations
that are better aligned with the decoder\textquoteright s representation
space. Consequently, the resampler can form effective prefix tokens
from the encoder features more easily, whereas the CLIP-ViT-B/32 text
encoder introduces a larger representation mismatch that the resampler
must learn to compensate for. Furthermore, we observe that Qwen3-Embedding-0.6B
reaches strong reconstruction performance earlier than Qwen3-0.6B,
which is likely because Qwen3-Embedding is explicitly optimized to
produce information dense embeddings for downstream use, providing
a more suitable input representation for reconstruction. These results
empirically justify the configuration of text auto-encoder models.

\subsection{Ablation Studies}

\subsubsection{Impact of Different Parameterizations \label{subsec:Impact-of-parameterization}}

Modaliff supports multiple parameterizations of the denoising model.
We study three diffusion parameterizations, including $\epsilon$-prediction,
$x_{0}$-prediction, and $v$-prediction, where $v=a_{t}\epsilon-\sigma_{t}x_{0}$.
We also include a Rectified Flow (RF) variant that uses the $\epsilon-x_{0}$
parameterization and the RF noise schedule~\cite{li2025omniflow,cha2026flowbind}.
For $\epsilon$-prediction, we normalize each modality using training
set mean and standard deviation, whereas $x_{0}$-prediction and $v$-prediction
are trained on raw latents. Moreover, we find that $\epsilon$-prediction
with a cosine schedule yields corrupted samples due to $SNR\left(T\right)\approx0$,
consistent with prior work~\cite{lin2024common}, so we instead use
a linear $\beta_{t}$ schedule and clip normalized values to {[}\textminus 8,8{]}
for stability. To save computational resources, we reduce the denoiser
to 159M parameters and evaluate both FID for T$\to$I and CIDEr for
I$\to$T on 5K MSCOCO validation samples.

As shown in Tab.~\ref{tab:param}, $x_{0}$-prediction achieves the
best performance on 4 out of 6 generation tasks, with the largest
improvement on T$\to$I and A$\to$I, while $\epsilon$-prediction
remains less stable and produces corrupted samples on image generation
tasks (see Fig.~\ref{fig:eps_failure}). We attribute this behavior
to the CLIP image latent space, where several dimensions remain heavy-tailed
and contain large-magnitude outliers even after per-modality standardization
(Fig.~\ref{fig:heavytail}). These latent statistics make $\epsilon$-prediction
difficult to learn, whereas $x_{0}$-prediction directly predicts
the clean latent and provides a more data-aligned objective. The mixed
noise-data parameterizations, including $v$-prediction and RF, underperform
$x_{0}$-prediction on most tasks. This difference is likely due to
the added complexity of mixed noise-data targets, which can be harder
to learn than direct $x_{0}$-prediction under the same model capacity,
consistent with findings in~\cite{li2025back}. These results support
our choice of $x_{0}$-prediction as a robust default for Modaliff
across modalities with diverse latent statistics.
\begin{center}
\begin{table}
\begin{centering}
\setlength{\tabcolsep}{1mm}%
\begin{tabular}{ccccccc}
\toprule 
\multirow{2}{*}{Prediction} & T$\to$I & I$\to$T & T$\to$A & A$\to$T & A$\to$I & I$\to$A\tabularnewline
\cmidrule{2-7}
 & FID$\downarrow$ & CIDEr$\uparrow$ & FAD$\downarrow$ & CIDEr$\uparrow$ & FID$\downarrow$ & FAD$\downarrow$\tabularnewline
\midrule
\midrule 
$\epsilon-x_{0}$ & 32.37 & \textbf{46.01} & 1.66 & \textbf{44.26} & 36.10 & 1.69\tabularnewline
\midrule 
$\epsilon$ & 74.29 & 22.31 & 1.82 & 41.29 & 159.46 & 1.91\tabularnewline
\midrule 
$a_{t}\epsilon-\sigma_{t}x_{0}$ & 28.39 & 29.57 & 1.74 & 41.08 & 34.27 & 1.87\tabularnewline
\midrule 
$x_{0}$ & \textbf{24.45} & 36.91 & \textbf{1.33} & 40.43 & \textbf{29.85} & \textbf{1.49}\tabularnewline
\bottomrule
\end{tabular}
\par\end{centering}
\caption{Fidelity metrics of Modaliff with different diffusion parameterizations.\label{tab:param}}
\end{table}
\begin{figure}
\centering{}%
\begin{minipage}[t]{0.4\textwidth}%
\begin{center}
\includegraphics[width=0.48\linewidth]{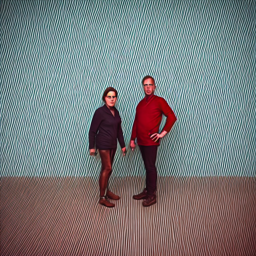}\hfill{}\includegraphics[width=0.48\linewidth]{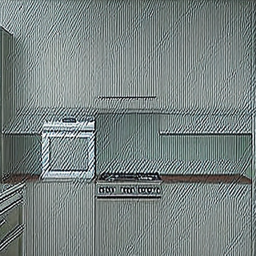}
\par\end{center}
\caption{\label{fig:eps_failure}Failure cases of $\epsilon$-prediction on
T$\to$I generation.}
\end{minipage}\hfill{}%
\begin{minipage}[t]{0.55\textwidth}%
\begin{center}
\includegraphics[width=0.48\linewidth]{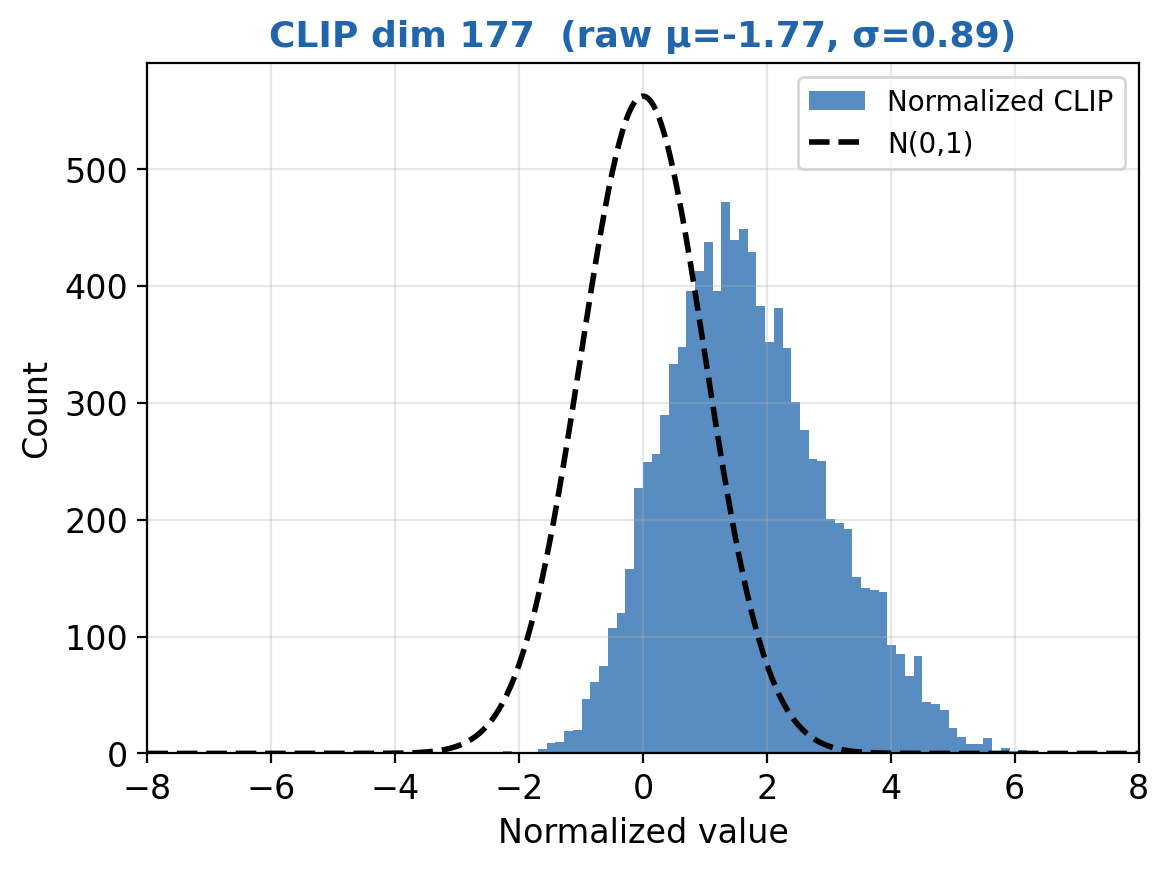}\hfill{}\includegraphics[width=0.48\linewidth]{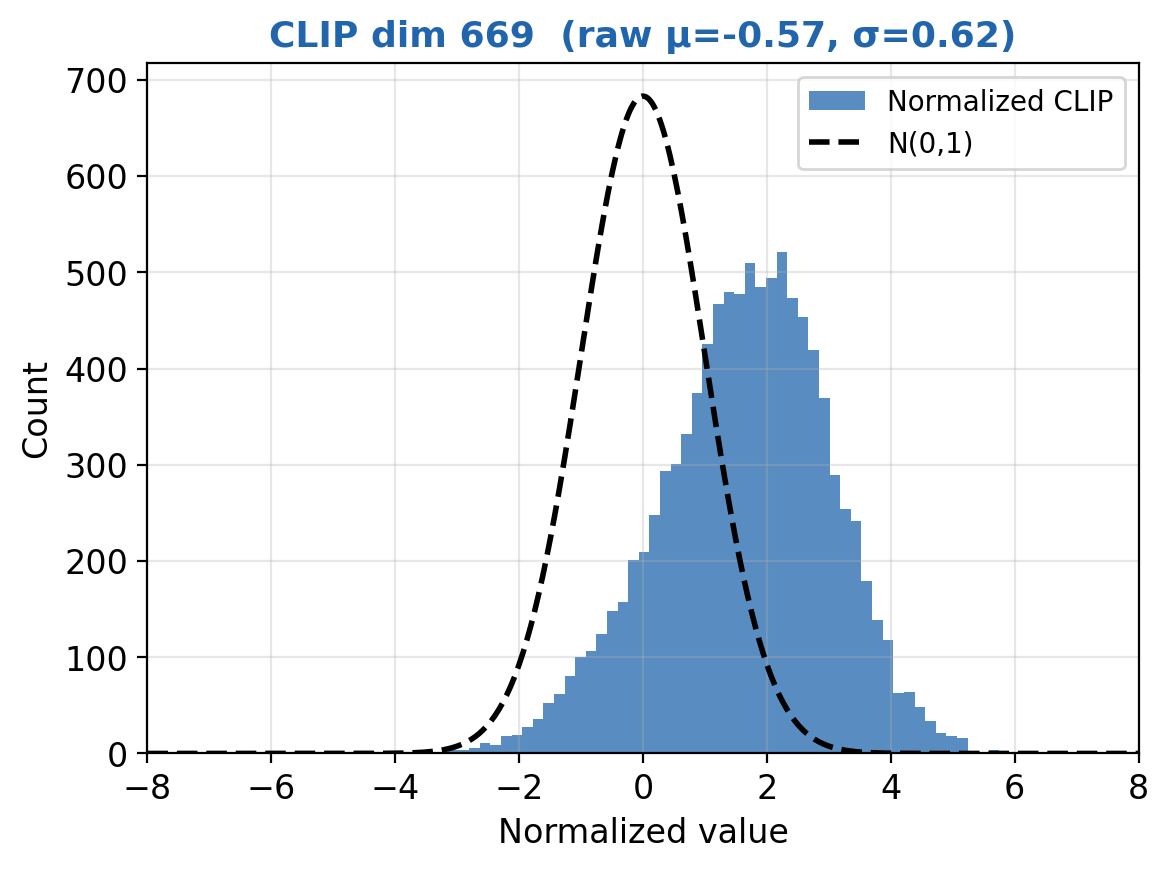}
\par\end{center}
\caption{\label{fig:heavytail}Heavy-tailed distributions in normalized CLIP
image latent dimensions.}
\end{minipage}
\end{figure}
\par\end{center}

\subsubsection{Impact of the Attribute Tuple and Token-wise AdaLN}

The attribute tuple $\left(r,i,t\right)$ and its token-wise adaptive
normalization are Modaliff's key mechanisms for resolving task ambiguity
in the unified sequence (Sec.~\ref{subsec:unified}). We isolate
their contribution with two ablated variants of the 159M model, evaluating
FID for T$\to$I and CIDEr for I$\to$T on 5K MSCOCO samples. \emph{W/o
attributes} removes the role and modality embeddings from both the
AdaLN conditioning and the token-content injection, reducing the per-token
conditioning vector to the timestep embedding, with $t=0$ for condition
tokens.\emph{ Seq.-level AdaLN} keeps the full attribute tuple but
replaces the per-token conditioning vectors with their unweighted
mean over the sequence, producing a single set of scale, shift, and
gating parameters broadcast to all tokens, following the conditioning
of DiT~\cite{peebles2023scalable}. 

\begin{figure}
\begin{centering}
\includegraphics[width=0.7\textheight]{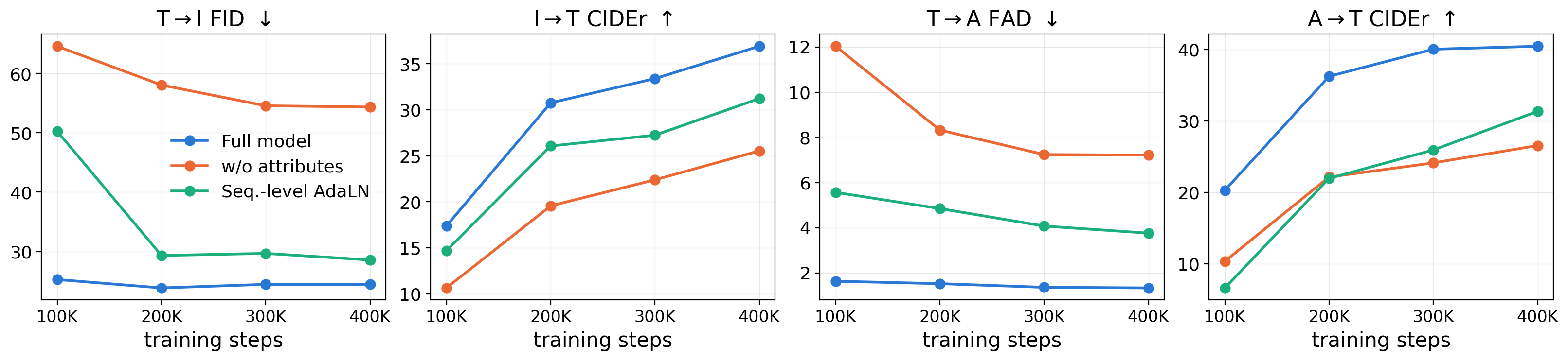}
\par\end{centering}
\caption{Learning curves of the architecture ablation on T$\to$I, I$\to$T,
T$\to$A, and A$\to$T.\label{fig:abla_curves_architecture}}

\end{figure}

Tab.~\ref{tab:arch_abl} reports the results after 400K iterations.
We observe that removing either mechanism degrades performance on
all reported tasks, and removing the attributes is the most damaging.
The collapse caused by removing the token attributes is expected from
the sequence layout, since T$\to$I, I$\to$T, T$\to$A, and A$\to$T
share the same number of tokens in the unified input sequence, making
it difficult for the shared model to distinguish between task configurations
and hindering the learning process, as reflected in Fig.~\ref{fig:abla_curves_architecture}.
Seq.-level AdaLN recovers much of this gap because the pooled vector
still identifies the task, but it remains behind the full model due
to a single set of modulation parameters to serve clean condition
tokens and noisy target tokens at once, limiting its ability to adapt
the computation across different generation configurations. These
results support the design of Modaliff, in which the attribute tuple
supplies the task information and token-wise AdaLN delivers it to
each token, allowing one shared denoiser to serve heterogeneous condition
and target layouts.
\begin{center}
\begin{table}
\begin{centering}
\setlength{\tabcolsep}{1mm}%
\begin{tabular}{ccccc}
\toprule 
\multirow{2}{*}{Variant} & T$\to$I & I$\to$T & T$\to$A & A$\to$T\tabularnewline
\cmidrule{2-5}
 & FID$\downarrow$ & CIDEr$\uparrow$ & FAD$\downarrow$ & CIDEr$\uparrow$\tabularnewline
\midrule
\midrule 
Full model & \textbf{24.45} & \textbf{36.91} & \textbf{1.33} & \textbf{40.43}\tabularnewline
\midrule 
w/o attributes & 54.35 & 25.53 & 7.21 & 26.54\tabularnewline
\midrule 
Seq.-level AdaLN & 28.57 & 31.22 & 3.76 & 31.32\tabularnewline
\bottomrule
\end{tabular}
\par\end{centering}
\caption{Ablation of the attribute tuple and token-wise adaptive normalization.\label{tab:arch_abl}}
\end{table}
\par\end{center}

\subsubsection{Impact of Guidance Scale for One-to-One Mappings}

\begin{figure}
\begin{centering}
\subfloat[Text-to-Image]{\begin{centering}
\includegraphics[width=0.23\textwidth]{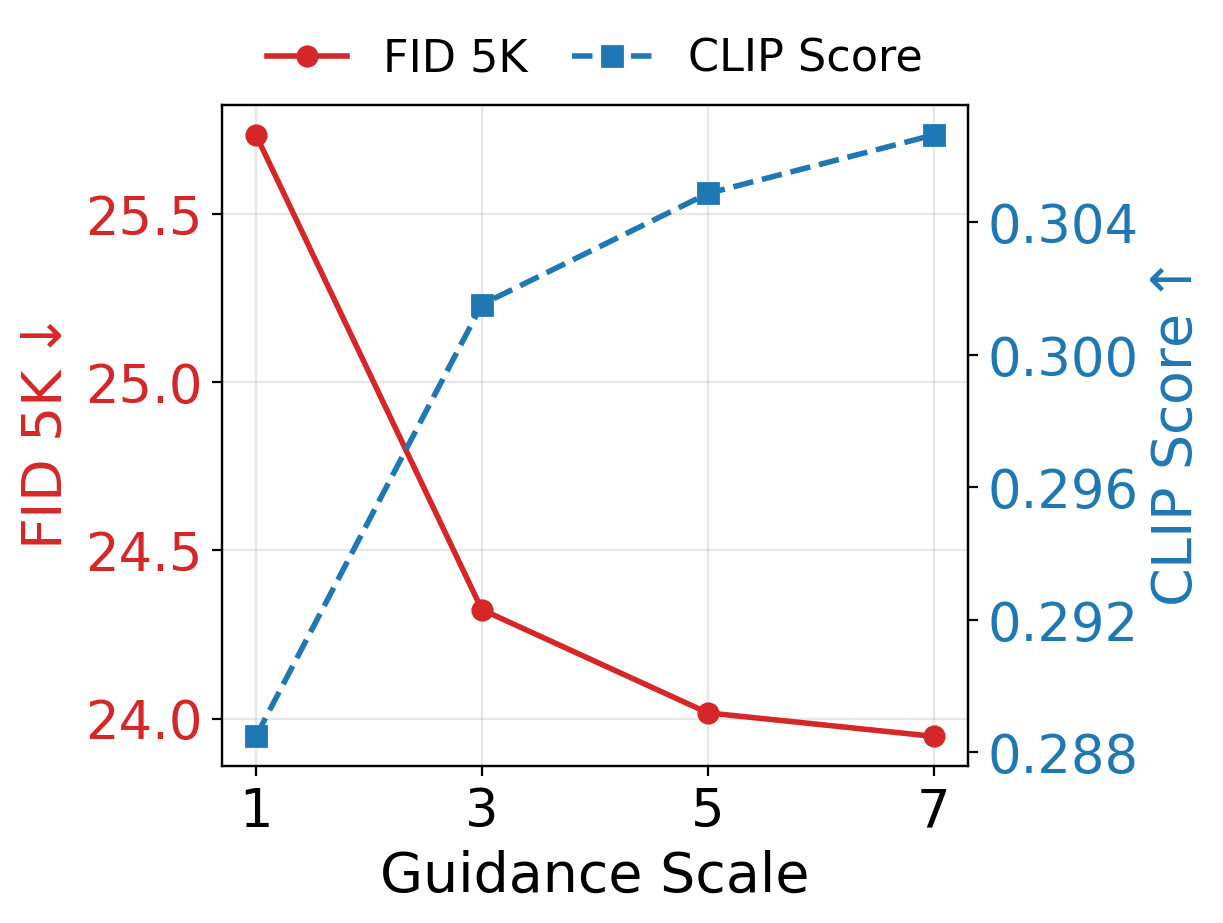}
\par\end{centering}
\centering{}}\subfloat[Image-to-Text]{\begin{centering}
\includegraphics[width=0.23\textwidth]{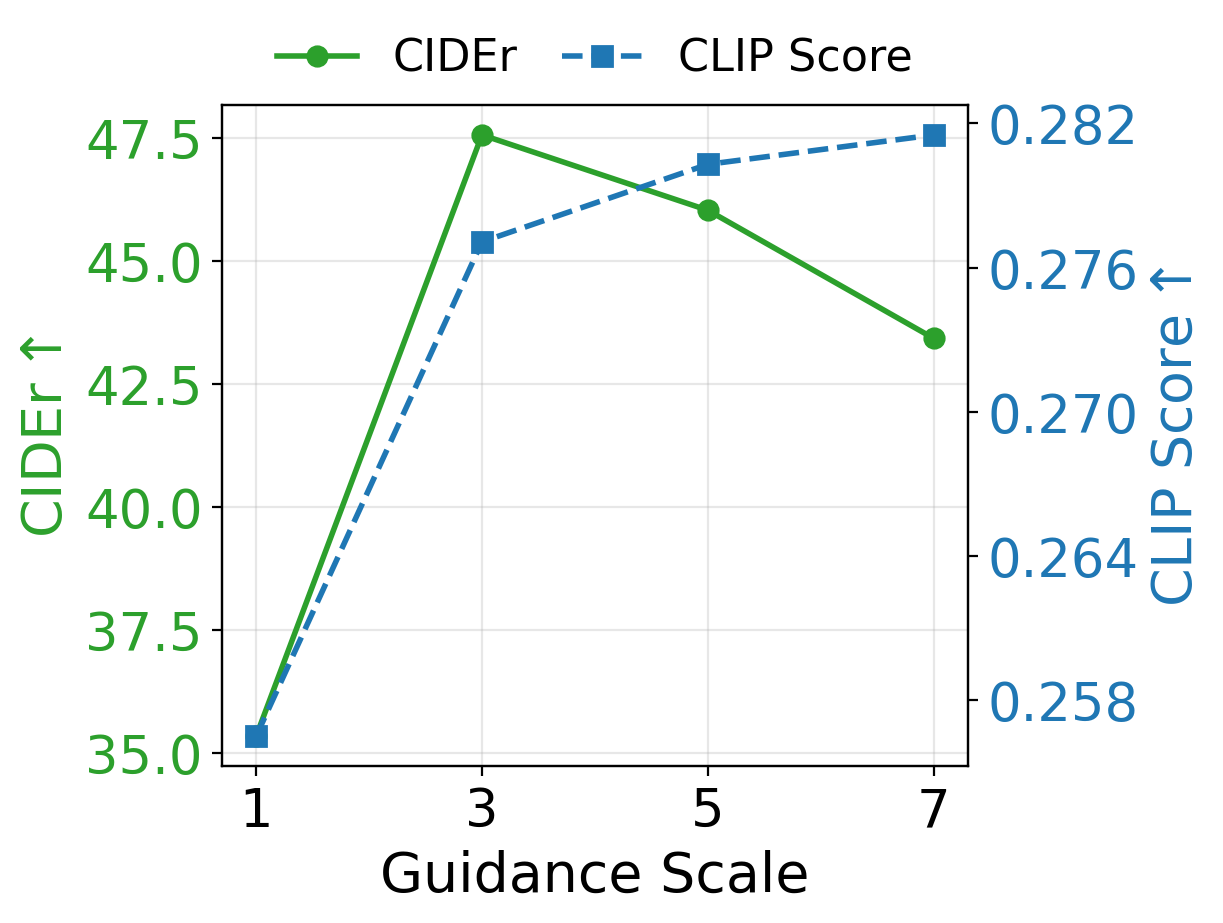}
\par\end{centering}
}\subfloat[Text-to-Audio]{\begin{centering}
\includegraphics[width=0.23\textwidth]{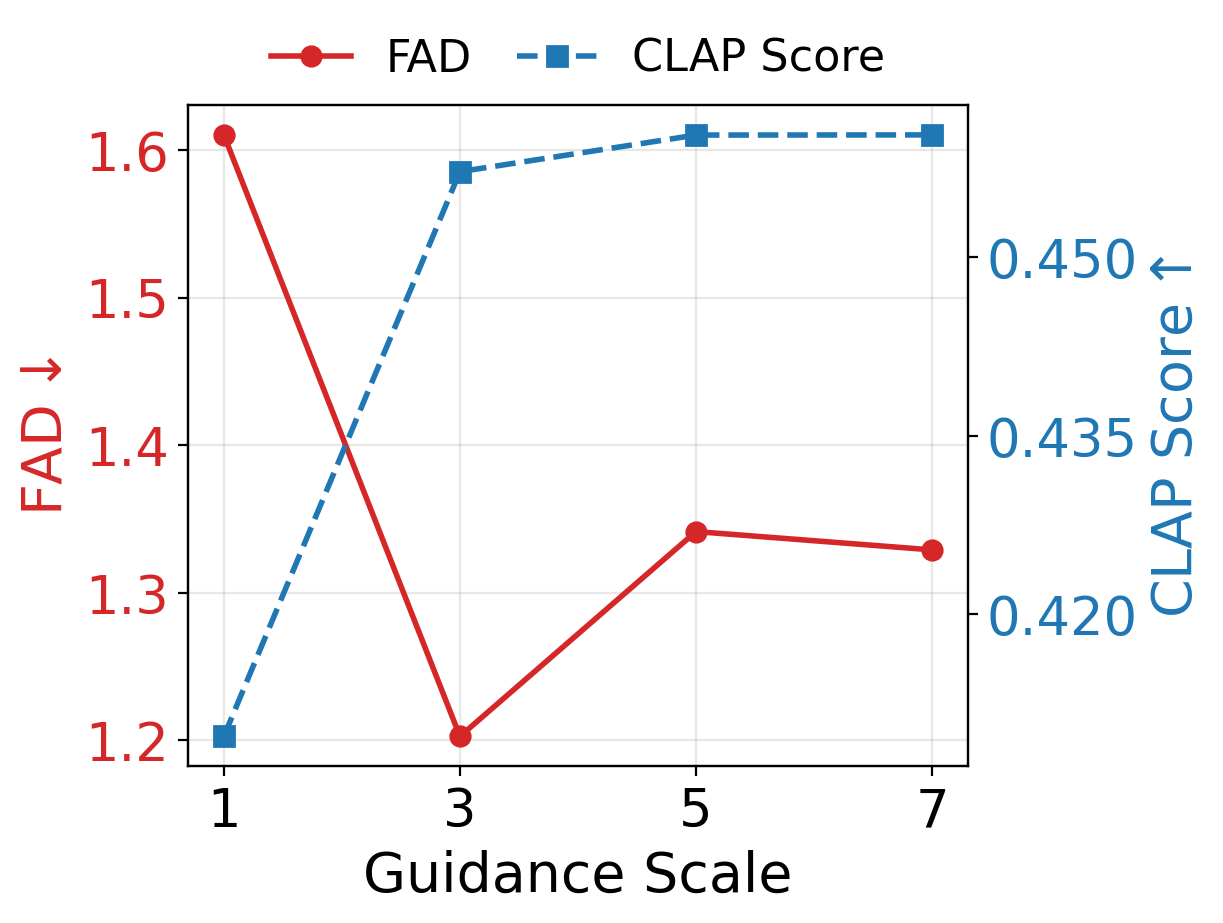}
\par\end{centering}
}\subfloat[Audio-to-Text]{\begin{centering}
\includegraphics[width=0.23\textwidth]{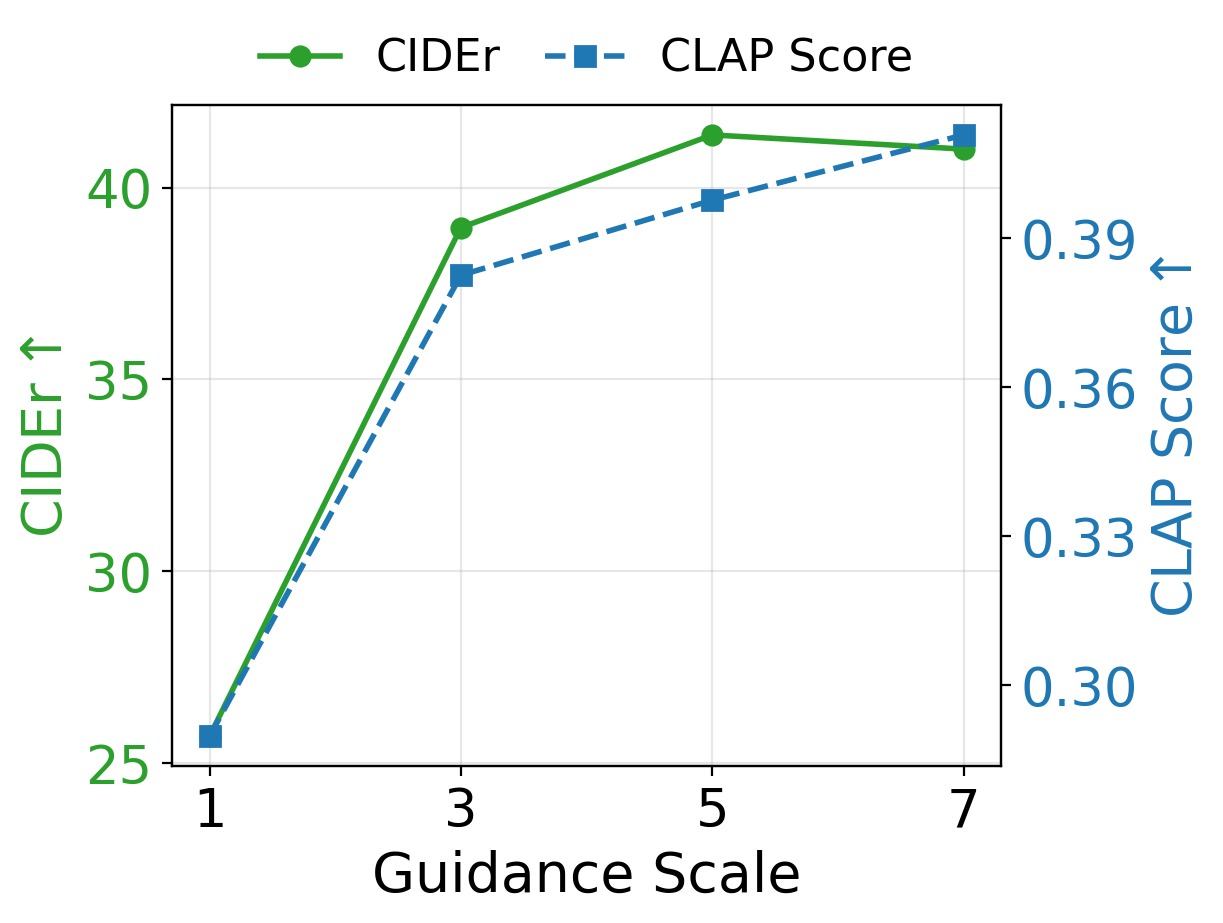}
\par\end{centering}
}
\par\end{centering}
\caption{Effect of classifier-free guidance scale on the quality and alignment
metrics across one-to-one generation tasks.\label{fig:abla_cfg}}
\end{figure}

We study the effect of classifier free guidance by varying the guidance
scale $w\in\left\{ 1,3,5,7\right\} $ and measuring both fidelity
and semantic alignment across one to one generation tasks. For evaluation
efficiency, we evaluate FID for T$\to$I and CIDEr for I$\to$T on
5K MSCOCO samples. Fig.~\ref{fig:abla_cfg} reports the corresponding
quality metrics and alignment scores for T$\to$I, I$\to$T, T$\to$A,
and A$\to$T. Across all directions, increasing $w$ consistently
improves alignment, as reflected by the rise of CLIP and CLAP scores,
indicating that stronger guidance enforces closer semantic agreement
with the conditioning signal. In contrast, fidelity follows a non
monotonic trend that reveals an over guidance regime. For T$\to$I
generation, FID improves as guidance increases from $w=3$ to $w=5$,
but at $w=7$ it begins to saturate, matching the trade-off pattern
observed in T$\to$A generation where FAD improves most at $w=3$
and then degrades at higher guidance. For captioning tasks, including
I$\to$T and A$\to$T, CIDEr improves sharply with $w$ and then saturates,
with a slight drop at $w=7$ even as alignment continues to improve.
Overall, these results indicate that moderate guidance offers the
best balance between fidelity and alignment across modalities, while
overly large guidance can over constrain sampling and reduce generation
quality.

\subsubsection{Sensitivity to the Composition Weights $w_{j}$ \label{subsec:wj_sweep}}

Eq.~\ref{eq:composed_score} assigns a guidance weight $w_{j}$ to
each condition, motivating an analysis of sensitivity to different
weight configurations. We evaluate (T+A)$\to$I by varying $w_{t}$
and $w_{a}$, and (A+I)$\to$T by varying $w_{i}$ and $w_{a}$. We
consider $w_{a}\in\left\{ 1,1.5,2,3\right\} $ and $w_{t},w_{i}\in\left\{ 2,3,4,5,6\right\} $,
yielding 20 configurations per task. Each configuration generates
1K test samples, which are evaluated using the ImageBind alignment
between the generated output and each condition.

\begin{figure}
\begin{centering}
\subfloat[(T+A)$\to$I]{\begin{centering}
\includegraphics[width=0.3\textwidth]{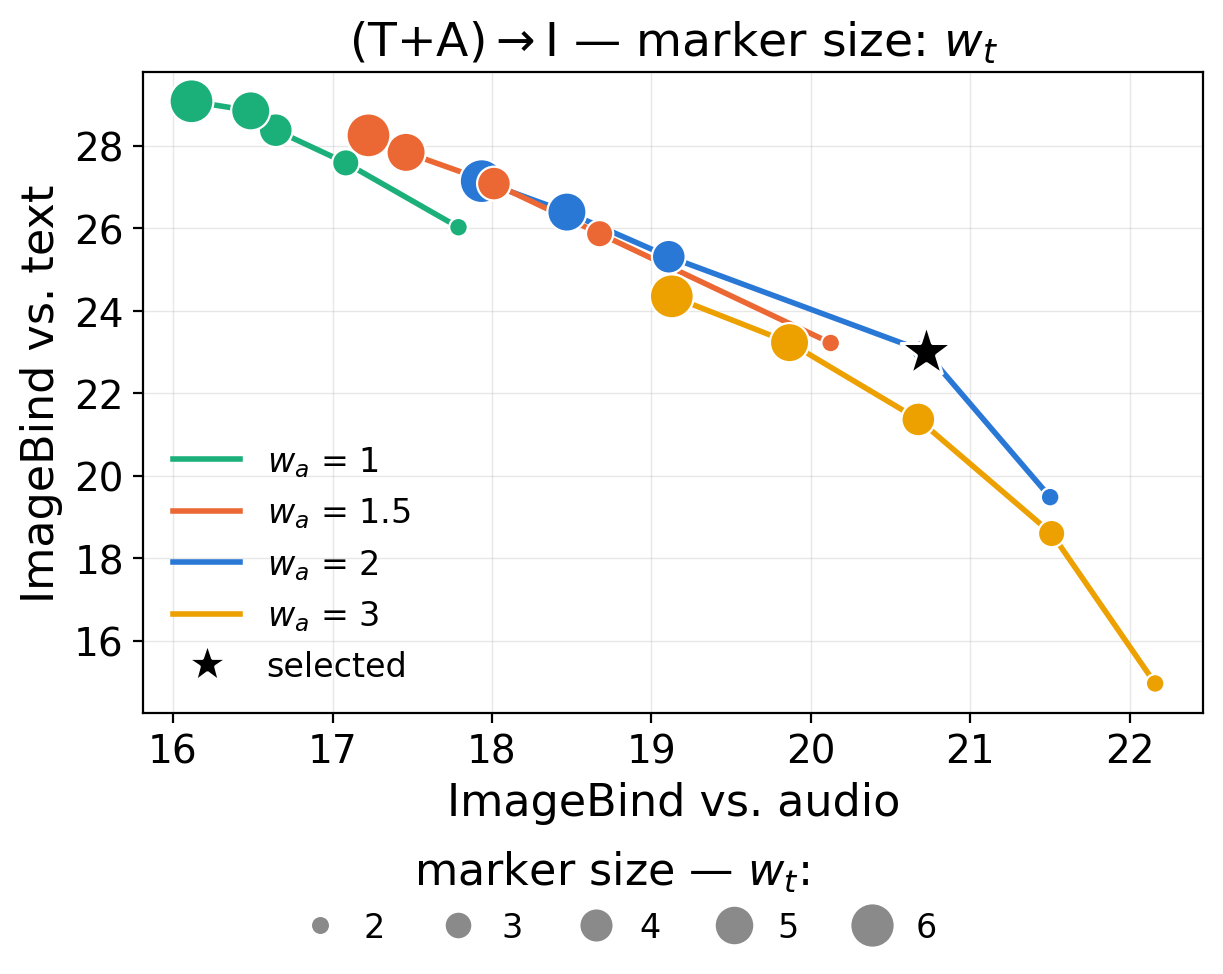}
\par\end{centering}
}\subfloat[(A+I)$\to$T]{\begin{centering}
\includegraphics[width=0.3\textwidth]{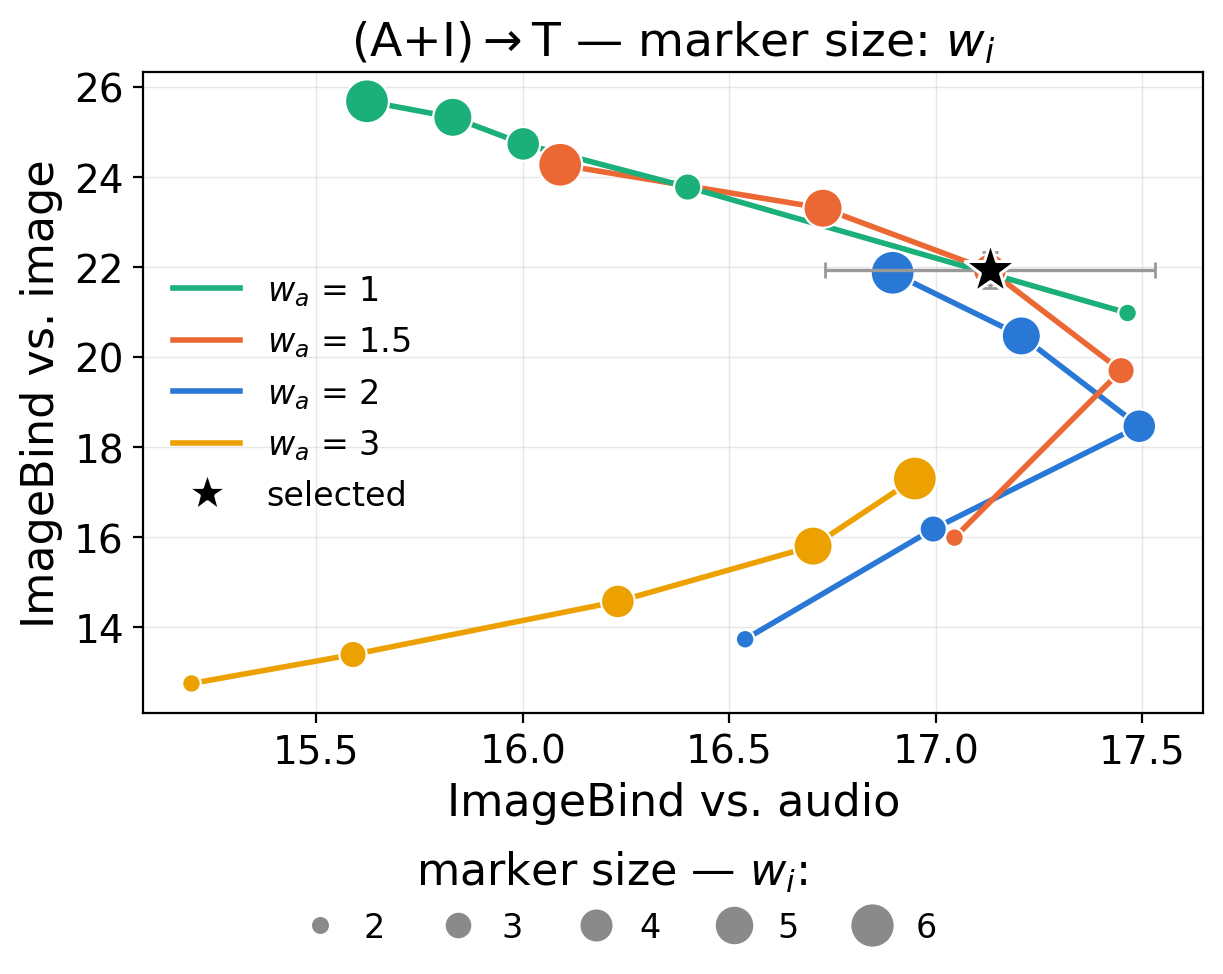}
\par\end{centering}

}\subfloat[Alignment scores for all configurations]{\begin{centering}
\includegraphics[width=0.36\textwidth]{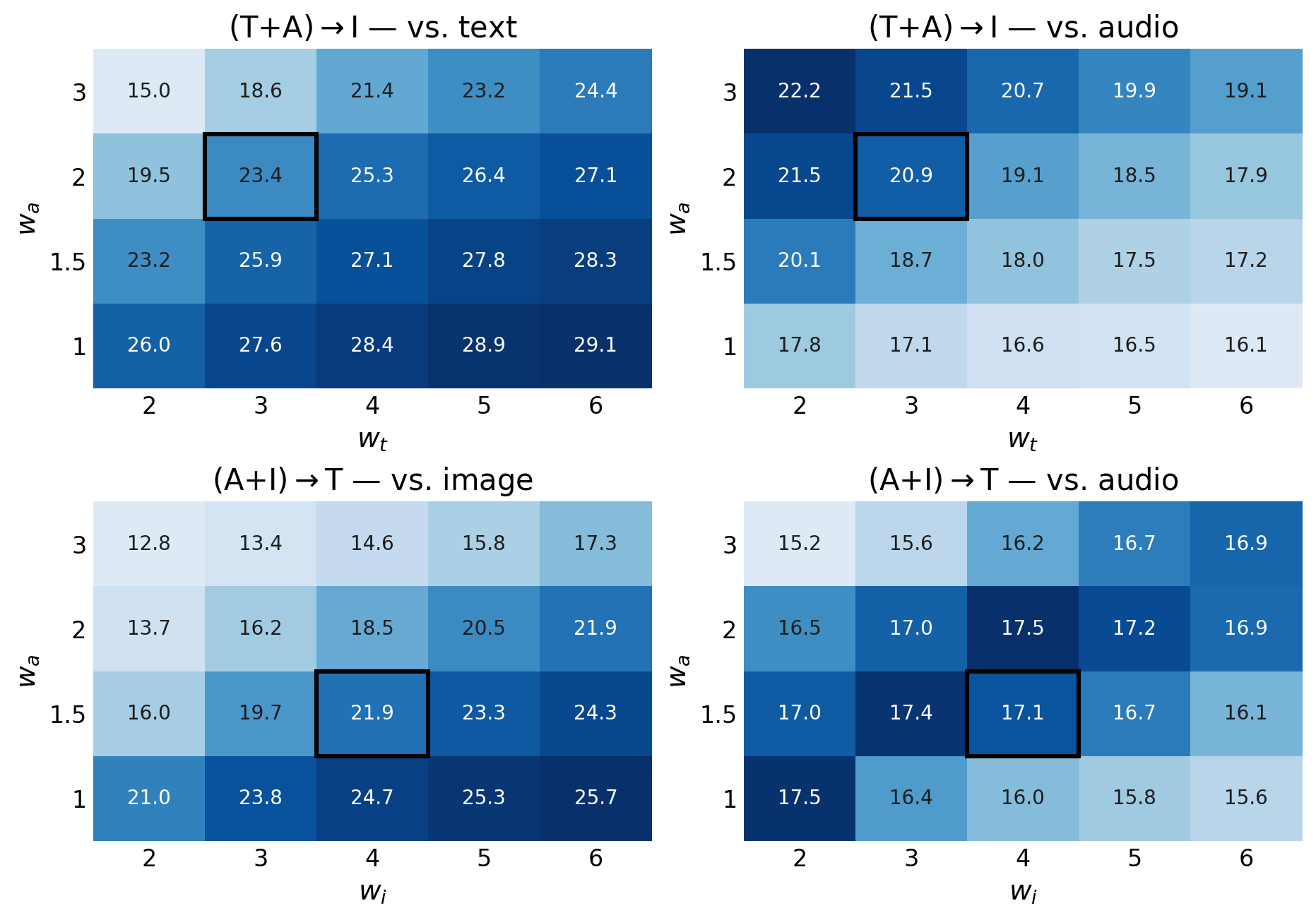}
\par\end{centering}
}
\par\end{centering}
\caption{Sensitivity to the composition weights on (T+A)$\to$I and (A+I)$\to$T.
In (a) and (b), each curve fixes $w_{a}$, marker size encodes the
weight of the other condition, and the star marks the selected configuration.
In (c), each cell reports the ImageBind alignment between the generated
output and one condition, with the selected configuration outlined
in black.\label{fig:wj_sweep}}

\end{figure}

Fig.~\ref{fig:wj_sweep} shows the task-dependent trade-off between
the two conditions. For (T+A)$\to$I, increasing $w_{t}$ improves
text alignment at the expense of audio alignment, while increasing
$w_{a}$ produces the opposite effect. Configurations with similar
ratios $w_{a}/w_{t}$ tend to occupy nearby regions of the trade-off
frontier, suggesting that relative weighting largely determines the
balance between the conditions. A similar trend is also reflected
in (A+I)$\to$T , where increasing $w_{i}$ generally favors image
alignment but degrades the audio alignment. Interestingly, for (A+I)$\to$T,
we observe that $w_{a}/w_{i}>0.5$ reduces alignment with both conditions
relative to the corresponding configuration with $w_{a}/w_{i}=0.5$,
further indicating that the relative guidance ratio largely controls
the balance between conditions and that large $w_{a}$ can push the
model into an over-guided regime that degrades overall caption quality.
Based on these results, we use $\left(w_{t},w_{a}\right)=\left(3,2\right)$
for (T+A)$\to$I and $\left(w_{a},w_{i}\right)=\left(1.5,4\right)$
for (A+I)$\to$T, marked by stars in Fig.~\ref{fig:wj_sweep}, keeping
alignment to both conditions competitive rather than maximizing either
alone. For (T+I)$\to$A, we keep the default $w_{t}=w_{i}=2$, which
already achieves competitive alignment to both conditions in Tab.~\ref{tab:multi_cond_results}.

\subsection{Qualitative Results}

\begin{figure}
\begin{centering}
\includegraphics[totalheight=0.9\textheight]{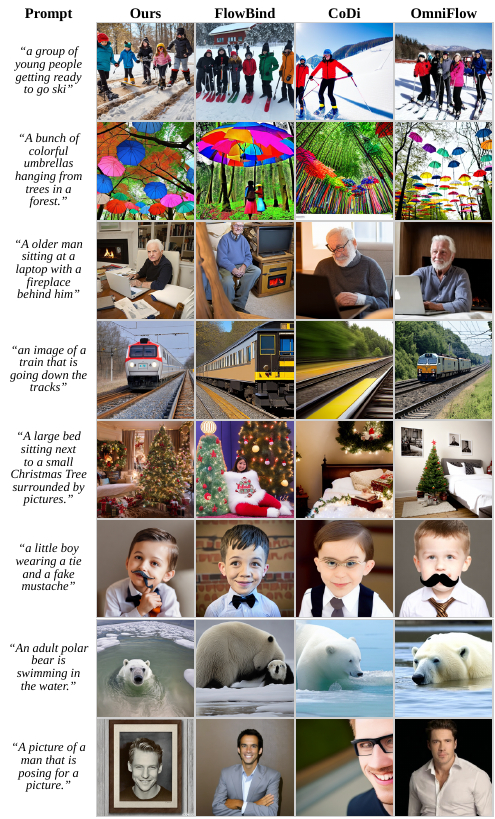}
\par\end{centering}
\caption{Qualitative results on T$\to$I generation.\label{fig:qual_t2i}}
\end{figure}

\begin{figure}
\begin{centering}
\includegraphics[width=0.7\textwidth]{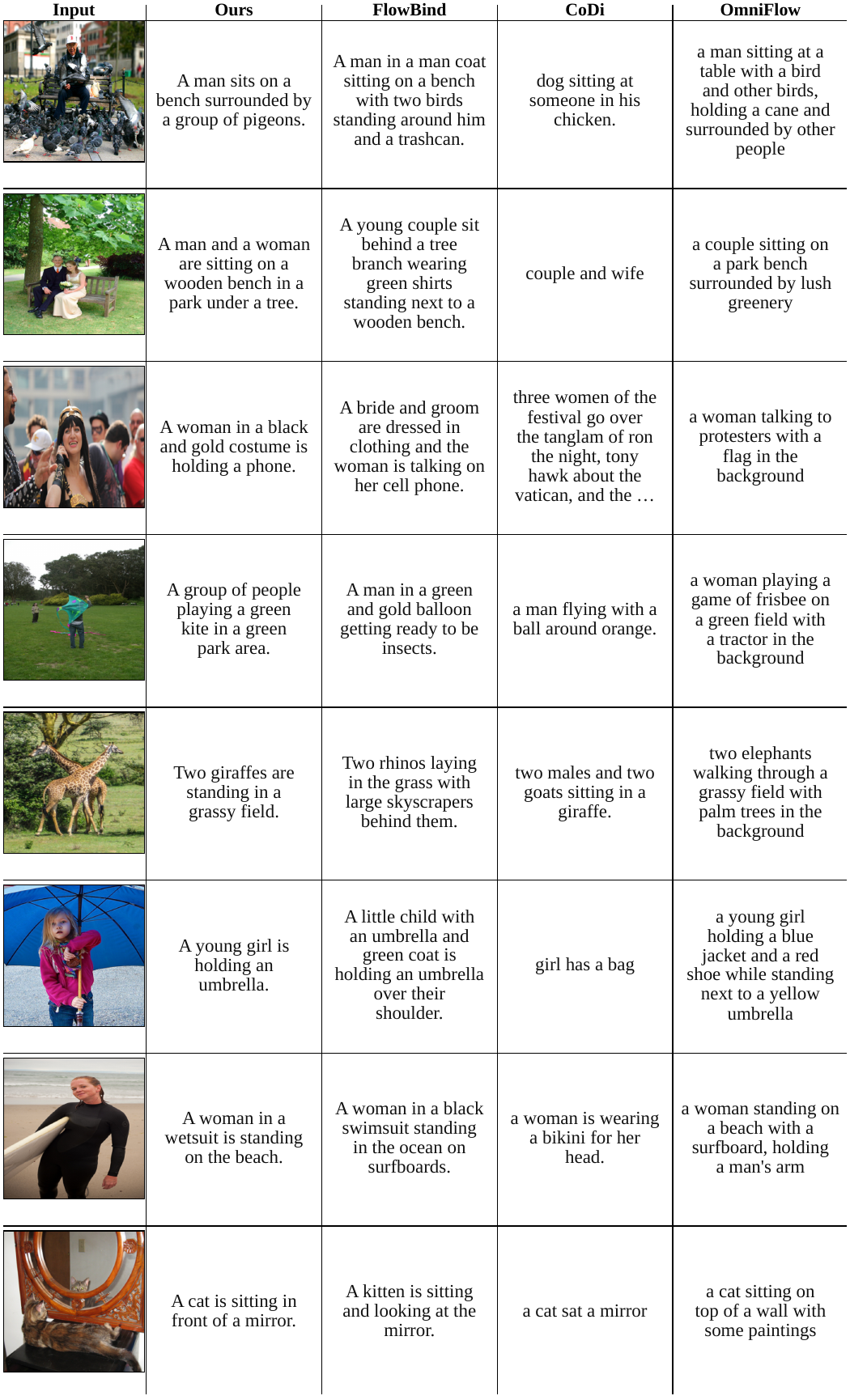}
\par\end{centering}
\caption{Qualitative results on I$\to$T generation.\label{fig:qual_i2t}}
\end{figure}

\begin{figure}
\begin{centering}
\includegraphics[width=0.7\textwidth]{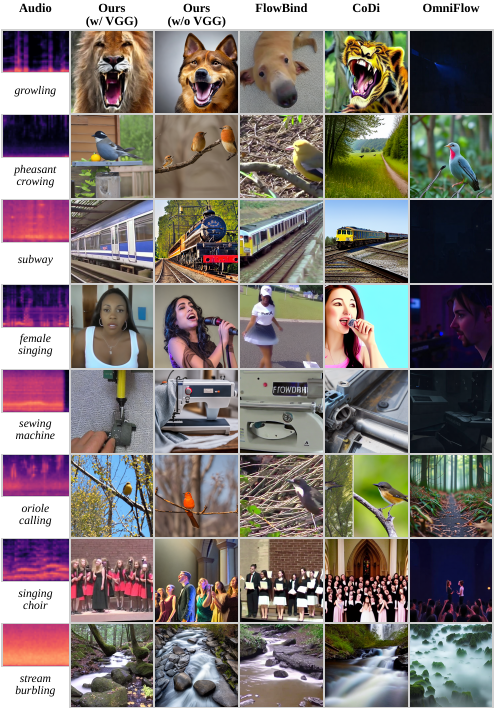}
\par\end{centering}
\caption{Qualitative results on A$\to$I generation.\label{fig:qual_a2i}}
\end{figure}

\begin{figure}
\begin{centering}
\includegraphics[width=0.7\textwidth]{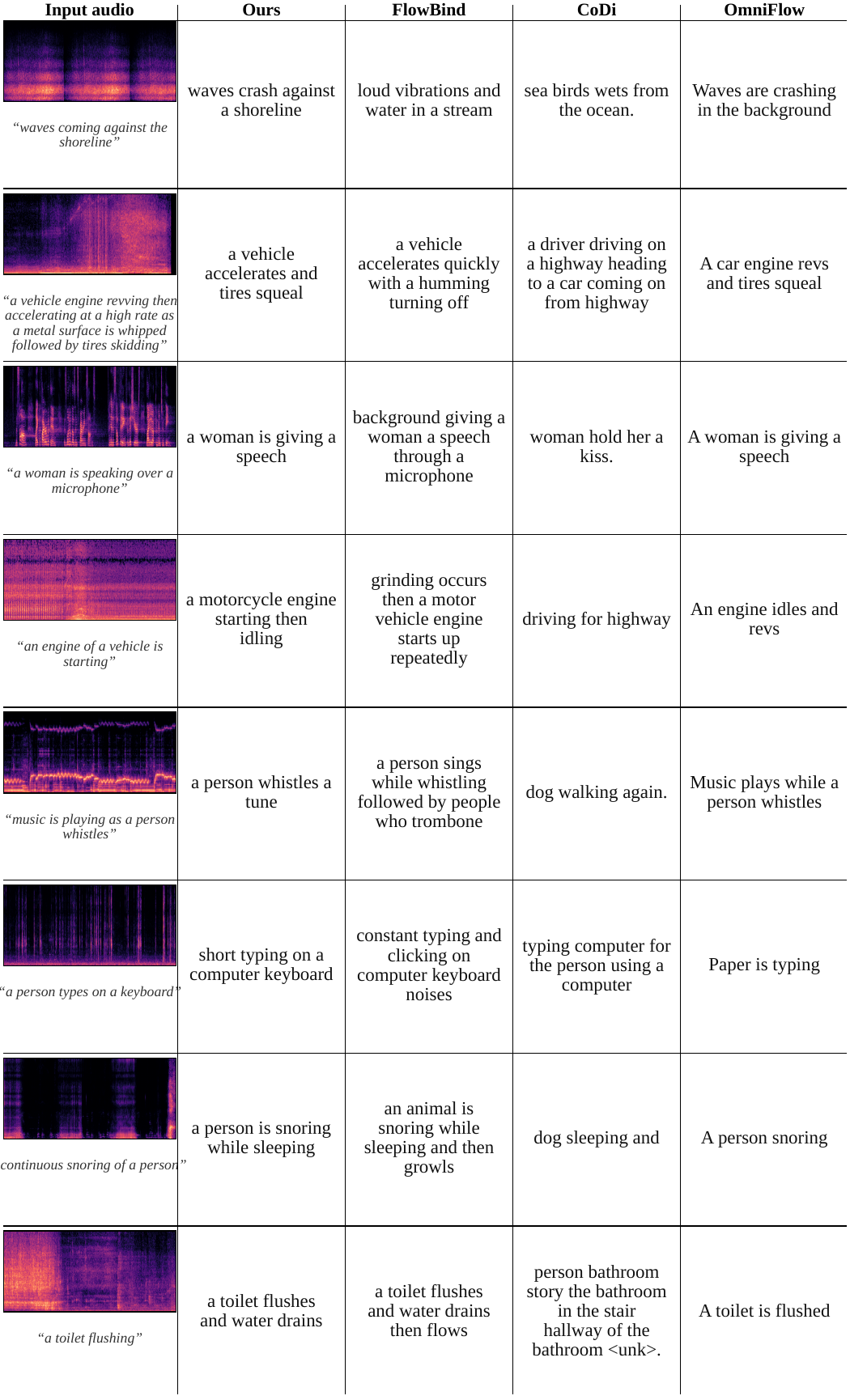}
\par\end{centering}
\caption{Qualitative results on A$\to$T generation.\label{fig:qual_a2t}}
\end{figure}

\begin{figure}
\begin{centering}
\includegraphics[width=0.7\textwidth]{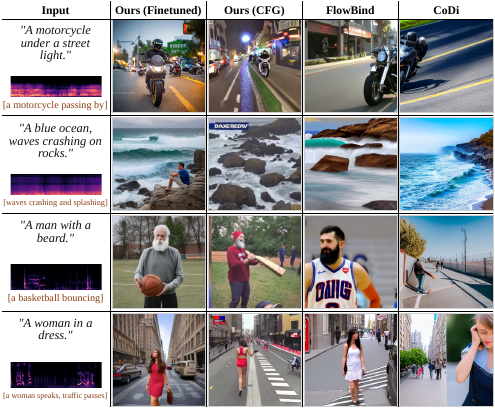}
\par\end{centering}
\caption{Qualitative results on (T + A)$\to$I generation.\label{fig:qual_ta2i}}
\end{figure}

\begin{figure}
\begin{centering}
\includegraphics[width=0.7\textwidth]{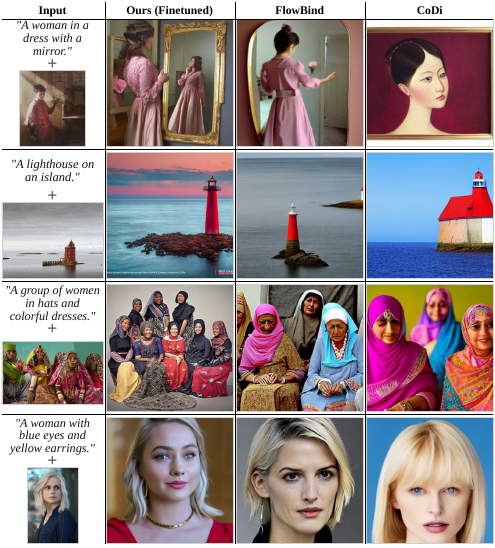}
\par\end{centering}
\caption{Qualitative results on (T + I)$\to$I generation.\label{fig:qual_ti2i}}
\end{figure}

\begin{figure}
\begin{centering}
\includegraphics[width=0.7\textwidth]{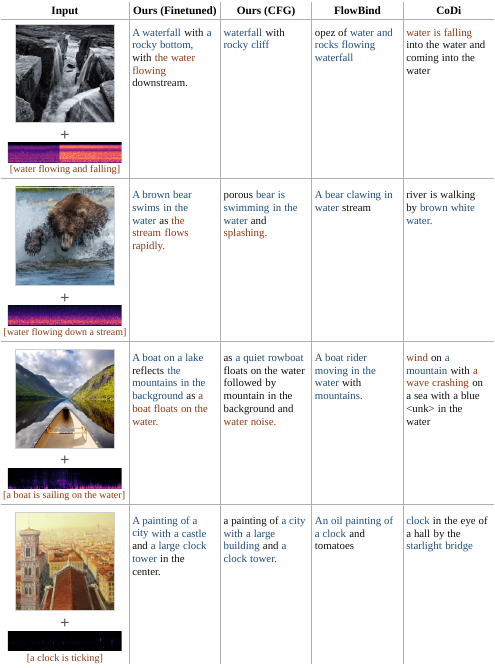}
\par\end{centering}
\caption{Qualitative results on (I + A)$\to$T generation.\label{fig:qual_ia2t}}
\end{figure}

We present qualitative results across a broad range of one-to-one
and multi-condition generation tasks in Figs.~\ref{fig:qual_t2i}--\ref{fig:qual_ia2t},
covering diverse people, objects, scenes, and semantic content. Modaliff
consistently produces high-quality outputs that remain faithful to
the provided conditions across text, image, and audio modalities.


